\documentclass{article} 
\usepackage{iclr2024_conference,times}

\usepackage{amsmath,amsfonts,bm}

\def\eqref#1{equation~\ref{#1}}

\def\1{\bm{1}}

\DeclareMathAlphabet{\mathsfit}{\encodingdefault}{\sfdefault}{m}{sl}
\SetMathAlphabet{\mathsfit}{bold}{\encodingdefault}{\sfdefault}{bx}{n}

\definecolor{darkgreen}{RGB}{47, 135, 91} 
\definecolor{orange}{RGB}{235, 122, 52} 
\definecolor{cyan}{RGB}{73, 167, 201}
\definecolor{darkcyan}{RGB}{67, 117, 148}
\definecolor{darkgray}{RGB}{116, 121, 125}

\usepackage[colorlinks=true,citecolor=darkcyan,
            linkcolor=red]{hyperref}%
\usepackage{url}
\usepackage{booktabs}
\usepackage{arydshln}
\usepackage{graphicx}
\usepackage{times}
\usepackage{colortbl}
\usepackage[normalem]{ulem}
\useunder{\uline}{\ul}{}
\usepackage{caption}
\usepackage{subcaption}
\usepackage[accsupp]{axessibility}  
\usepackage{stfloats}
\usepackage{bm}
\usepackage{flushend}
\usepackage{lipsum}
\usepackage{wrapfig}
\usepackage{overpic}
\usepackage[ruled,noend,vlined]{algorithm2e}
\usepackage{xfrac}          
\usepackage{setspace}
\usepackage{pifont}
\newcommand{\cmark}{\ding{51}}%
\usepackage{tikz}
\usepackage{comment}
\usepackage[scr=boondoxo]{mathalpha}
\makeatletter
\def\adl@drawiv#1#2#3{%
        \hskip.5\tabcolsep
        \xleaders#3{#2.5\@tempdimb #1{1}#2.5\@tempdimb}%
                #2\z@ plus1fil minus1fil\relax
        \hskip.5\tabcolsep}
\newcommand{\cdashlinelr}[1]{%
  \noalign{\vskip\aboverulesep
           \global\let\@dashdrawstore\adl@draw
           \global\let\adl@draw\adl@drawiv}
  \cdashline{#1}
  \noalign{\global\let\adl@draw\@dashdrawstore
           \vskip\belowrulesep}}
\makeatother
\newcommand{\MYhref}[3][blue]{\href{#2}{\color{#1}{#3}}}%

\title{BECLR: Batch Enhanced Contrastive\\ Few-Shot Learning}

\author{Stylianos Poulakakis-Daktylidis$^1$ \& Hadi Jamali-Rad$^{1,2}$\\
$^1$Delft University of Technology (TU Delft), The Netherlands\\
$^2$Shell Global Solutions International B.V., Amsterdam, The Netherlands \\
\texttt{\{s.p.mrpoulakakis-daktylidis, h.jamalirad\}@tudelft.nl}
}

\newcommand{\ourmethod}{\texttt{BECLR}}
\newcommand{\ourmodule}{\texttt{DyCE}}
\newcommand{\ourft}{\texttt{OpTA}}

\iclrfinalcopy 
\begin{document}

\maketitle

\vspace{-0.2cm}
\begin{abstract}
\vspace{-0.2cm}
Learning quickly from very few labeled samples is a fundamental attribute that separates machines and humans in the era of deep representation learning. Unsupervised few-shot learning (U-FSL) aspires to bridge this gap by discarding the reliance on annotations at training time. Intrigued by the success of contrastive learning approaches in the realm of U-FSL, we structurally approach their shortcomings in both pretraining and downstream inference stages. We propose a novel \texttt{Dy}namic \texttt{C}lustered m\texttt{E}mory (\ourmodule{}) module to promote a highly separable latent representation space for \emph{enhancing positive sampling} at the pretraining phase and infusing implicit class-level insights into unsupervised contrastive learning. We then tackle the, somehow overlooked yet critical, issue of \emph{sample bias} at the few-shot inference stage. We propose an iterative \texttt{Op}timal \texttt{T}ransport-based distribution \texttt{A}lignment (\ourft{}) strategy and demonstrate that it efficiently addresses the problem, especially in low-shot scenarios where FSL approaches suffer the most from sample bias. We later on discuss that \ourmodule{} and \ourft{} are two intertwined pieces of a novel end-to-end approach (we coin as \ourmethod{}), constructively magnifying each other's impact. We then present a suite of extensive quantitative and qualitative experimentation to corroborate that \ourmethod{} sets a new state-of-the-art across ALL existing U-FSL benchmarks (to the best of our knowledge), and significantly outperforms the best of the current baselines (codebase available at \MYhref[darkcyan]{https://github.com/stypoumic/BECLR}{GitHub}).

\end{abstract}

\vspace{-0.3cm}
\section{Introduction}
\label{sec:intro}
\vspace{-0.2cm}

\begin{wrapfigure}{r}{0.47\columnwidth}
    \vspace{-10pt}
    \begin{overpic}[abs, unit=1cm, width=0.47\columnwidth, clip, percent]{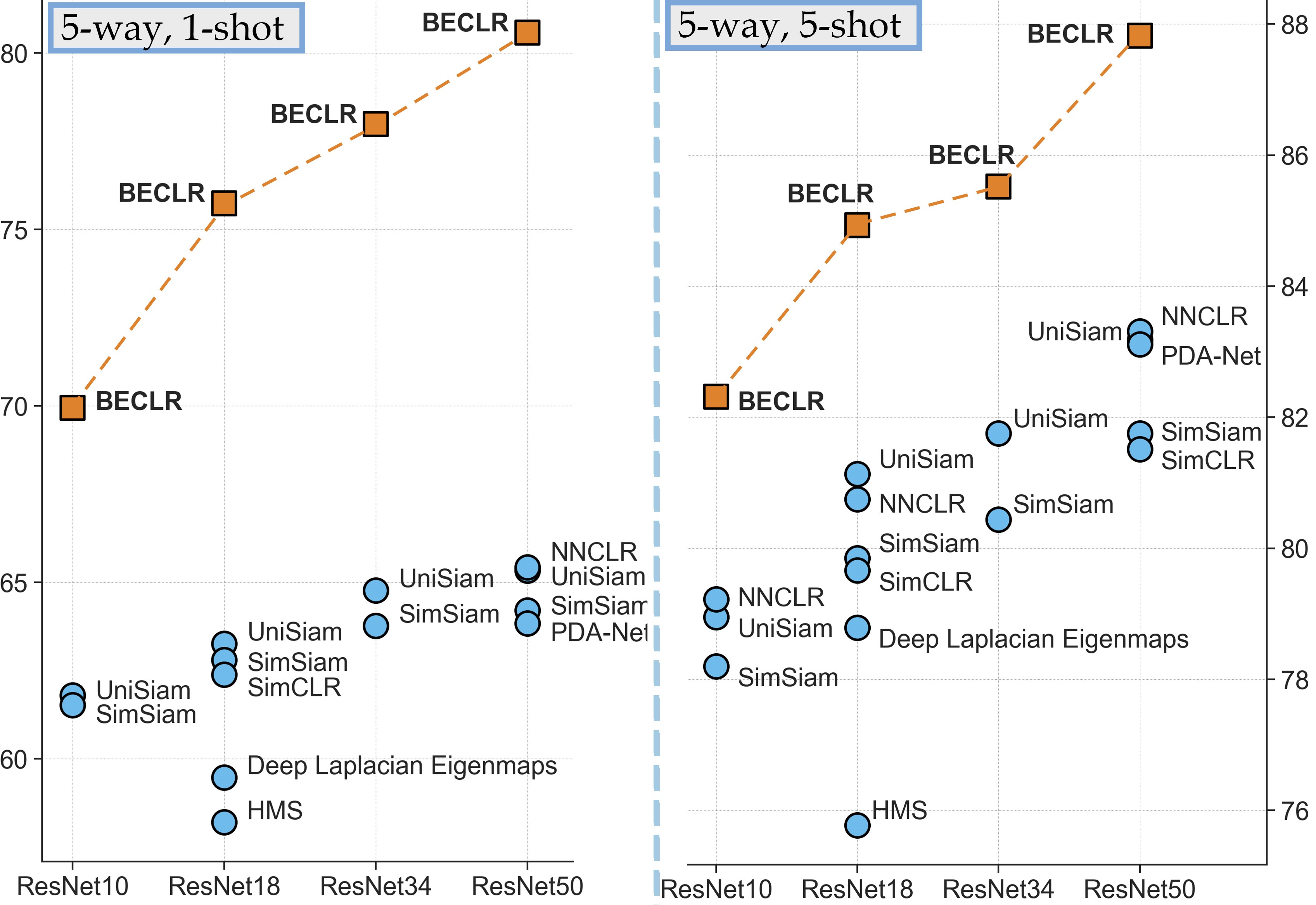}
    \end{overpic}
    \caption{\small miniImageNet ($5$-way, $1$-shot, left) and ($5$-way, $5$-shot, right) accuracy in the U-FSL landscape. \ourmethod{} sets a new state-of-the-art in all settings by a significant margin.}
    \label{fig:teaser}
    \vspace{-8pt}
\end{wrapfigure}
Achieving acceptable performance in deep representation learning comes at the cost of humongous data collection, laborious annotation, and excessive supervision. As we move towards more complex downstream tasks, this becomes increasingly prohibitive; in other words, supervised representation learning simply does not scale. In stark contrast, humans can quickly learn new tasks from a handful of samples, without extensive supervision. Few-shot learning (FSL) aspires to bridge this fundamental gap between humans and machines, using a suite of approaches such as metric learning \citep{wang2019simpleshot,bateni2020improved,yang2020dpgn}, meta-learning \citep{maml,rajeswaran2019meta,LEO}, and probabilistic learning \citep{iakovleva2020meta,hu2020empirical,zhang2021shallow}. FSL has shown promising results in a supervised setting so far on a number of benchmarks \citep{hu2022pushing,singh2022transductive,hu2023adaptive}; however, the need for supervision still lingers on. This has led to the emergence of a new divide called unsupervised FSL (U-FSL). The stages of U-FSL are the same as their supervised counterparts: pretraining on a large dataset of \emph{base} classes followed by fast adaptation and inference to unseen few-shot tasks (of \emph{novel} classes). The extra challenge here is the absence of labels during pretraining. U-FSL approaches have gained an upsurge of attention most recently owing to their practical significance and close ties to self-supervised learning. 

The goal of pretraining in U-FSL is to learn a feature extractor (a.k.a encoder) to capture the global structure of the unlabeled data. This is followed by fast adaptation of the frozen encoder to unseen tasks typically through a simple linear classifier (e.g., a logistic regression classifier). The body of literature here can be summarized into two main categories: (i) meta-learning-based pretraining where synthetic few-shot tasks resembling the downstream inference are generated for episodic training of the encoder \citep{Cactus-proto,umtra,khodadadeh2020unsupervised}; (ii) (non-episodic) transfer-learning approaches where pretraining boils down to learning optimal representations suitable for downstream few-shot tasks \citep{prototransfer,ubc-fsl,deepeigenmaps,psco}. Recent studies demonstrate that (more) complex meta-learning approaches are data-inefficient \citep{dhillon2019baseline,tian2020rethinking}, and that the transfer-learning-based methods outperform their meta-learning counterparts. More specifically, the state-of-the-art in this space is currently occupied by approaches based on \emph{contrastive learning}, from the transfer-learning divide, achieving top performance across a wide variety of benchmarks \citep{pdanet,unisiam}. The underlying idea of contrastive representation learning \citep{simclr,moco} is to attract ``positive'' samples in the representation space while repelling ``negative'' ones. To efficiently materialize this, some contrastive learning approaches incorporate memory queues to alleviate the need for larger batch sizes \citep{zhuang2019local,moco,nnclr,psco}. 

\textbf{Key Idea I:} \textit{Going beyond instance-level contrastive learning.} Operating under the unsupervised setting, contrastive FSL approaches typically enforce consistency only at the \emph{instance level}, where each image within the batch and its augmentations correspond to a unique class, which is an unrealistic but seemingly unavoidable assumption. The pitfall here is that potential positive samples present within the same batch might then be repelled in the representation space, hampering the overall performance. We argue that infusing a semblance of class (or membership)-level insights into the unsupervised contrastive paradigm is essential. Our key idea to address this is extending the concept of memory queues by introducing inherent membership clusters represented by dynamically updated prototypes, while circumventing the need for large batch sizes. This enables the proposed pretraining approach to sample more meaningful positive pairs owing to a novel \texttt{Dy}namic \texttt{C}lustered m\texttt{E}mory (\ourmodule{}) module. While maintaining a fixed memory size (same as queues), \ourmodule{} efficiently constructs and dynamically updates separable memory clusters.     

\textbf{Key Idea II:} \textit{Addressing inherent sample bias in FSL.} The base (pretraining) and novel (inference) classes are either mutually exclusive classes of the same dataset or originate from different datasets - both scenarios are investigated in this paper (in Section~\ref{sec:experimental-eval}). 
This distribution shift poses a significant challenge at inference time for the swift adaptation to the novel classes. This is further aggravated due to access to only a few labeled (a.k.a \emph{support}) samples within the few-shot task because the support samples are typically not representative of the larger unlabeled (a.k.a \emph{query}) set. We refer to this phenomenon as \emph{sample bias}, highlighting that it is overlooked by most (U-)FSL baselines. To address this issue, we introduce an \texttt{Op}timal \texttt{T}ransport-based distribution \texttt{A}lignment (\ourft{}) add-on module within the supervised inference step. \ourft{} imposes no additional learnable parameters, yet efficiently aligns the representations of the labeled support and the unlabeled query sets, right before the final supervised inference step. Later on in Section~\ref{sec:experimental-eval}, we demonstrate that these two novel modules (\ourmodule{} and \ourft{}) are actually intertwined and amplify one another. Combining these two key ideas, we propose an end-to-end U-FSL approach coined as \texttt{B}atch-\texttt{E}nhanced \texttt{C}ontrastive \texttt{L}ea\texttt{R}ning (\ourmethod{}). Our main contributions can be summarized as follows:
\begin{enumerate}
\item[I.] We introduce \ourmethod{} to structurally address two key shortcomings of the prior art in U-FSL at pretraining and inference stages. At pretraining, we propose to infuse implicit class-level insights into the contrastive learning framework through a novel dynamic clustered memory (coined as \ourmodule{}). Iterative updates through \ourmodule{} help establish a highly separable partitioned latent space, which in turn promotes more meaningful positive sampling.
\item[II.] We then articulate and address the inherent sample bias in (U-)FSL through a novel add-on module (coined as \ourft{}) at the inference stage of \ourmethod{}. We show that this strategy helps mitigate the distribution shift between query and support sets. This manifests its significant impact in low-shot scenarios where FSL approaches suffer the most from sample bias. 
%

\item[III.] We perform \emph{extensive} experimentation to demonstrate that \ourmethod{} sets a new state-of-the-art in ALL established U-FSL benchmarks; e.g. miniImageNet (see Fig.~\ref{fig:teaser}), tieredImageNet, CIFAR-FS, FC$100$, outperforming ALL existing baselines by a significant margin (up to $14\%$, $12\%$, $15\%$, $5.5\%$ in $1$-shot settings, respectively), to the best of our knowledge.
\end{enumerate}

\vspace{-0.3cm}
\section{Related Work}
\label{sec:rel-work}
\vspace{-0.3cm}
\textbf{Self-Supervised Learning (SSL).}\label{sssec:ssl-rel} 
It has been approached from a variety of perspectives \citep{balestriero2023cookbook}. Deep metric learning methods \citep{simclr,moco,SwAV,nnclr}, build on the principle of a contrastive loss and encourage similarity between semantically transformed views of an image. Redundancy reduction approaches \citep{barlowtwins,vicreg} infer the relationship between inputs by analyzing their cross-covariance matrices. Self-distillation methods \citep{byol,simsiam,oquab2023dinov2} pass different views to two separate encoders and map one to the other via a predictor. Most of these approaches construct a contrastive setting, where a symmetric or (more commonly) asymmetric Siamese network \citep{koch2015siamese} is trained with a variant of the infoNCE\citep{infonce} loss.

\textbf{Unsupervised Few-Shot Learning (U-FSL).}\label{sssec:ufsl-rel} The objective here is to pretrain a model from a large \emph{unlabeled} dataset of base classes, akin to SSL, but tailored so that it can quickly generalize to unseen downstream FSL tasks. Meta-learning approaches \citep{meta_gmvae,hms} generate synthetic learning episodes for pretraining, which mimic downstream FSL tasks. Here, PsCo \citep{psco} utilizes a student-teacher momentum network and optimal transport for creating pseudo-supervised episodes from a memory queue. Despite its elegant form, meta-learning has been shown to be data-inefficient in U-FSL \citep{dhillon2019baseline,tian2020rethinking}. On the other hand, transfer-learning approaches \citep{lf2cs, antoniou2019assume, wang2022contrastive}, follow a simpler non-episodic pretraining, focused on representation quality. Notably, contrastive learning methods, such as PDA-Net \citep{pdanet} and UniSiam \citep{unisiam}, currently hold the state-of-the-art. Our proposed approach also operates within the contrastive learning premise, but also employs a dynamic clustered memory module (\ourmodule{}) for infusing membership/class-level insights within the instance-level contrastive framework. Here, SAMPTransfer \citep{samptransfer} takes a different path and tries to ingrain \emph{implicit} global membership-level insights through message passing on a graph neural network; however, the computational burden of this approach significantly hampers its performance with (and scale-up to) deeper backbones than \texttt{Conv4}.

\textbf{Sample Bias in (U-)FSL.} Part of the challenge in (U-)FSL lies in the domain difference between base and novel classes. To make matters worse, estimating class distributions only from a few support samples is inherently biased, which we refer to as \emph{sample bias}. To address sample bias, \citet{pdanet} propose to enhance the support set with additional base-class images, \citet{xu2022alleviating} project support samples farther from the task centroid, while \citet{yang2021free} use a calibrated distribution for drawing more support samples, yet all these methods are dependent on base-class characteristics. On the other hand, \citet{ghaffari2021importance,wang2022revisit} utilize Firth bias reduction to alleviate the bias in the logistic classifier itself, yet are prone to over-fitting. In contrast, the proposed \ourft{} module requires no fine-tuning and does not depend on the pretraining dataset.



\vspace{-0.1cm}
\section{Problem Statement: Unsupervised Few-Shot Learning}
\label{sec:prelim}
\vspace{-0.3cm}
We follow the most commonly adopted setting in the literature \citep{pdanet, ubc-fsl, unisiam, psco}, which consists of: an unsupervised pretraining, followed by a supervised inference (a.k.a fine-tuning) strategy. Formally, we consider a large unlabeled dataset $\mathcal{D}_{\textup{tr}} = \{{\bm x}_i\}$ of so-called ``base'' classes for pretraining the model. The inference phase then involves transferring the model to unseen few-shot downstream tasks $\mathcal{T}_i$, drawn from a smaller labeled test dataset of so-called ``novel'' classes $\mathcal{D}_{\textup{tst}} = \{({\bm x}_i, y_i)\}$, with $y_i$ denoting the label of sample $\bm{x}_i$. Each task $\mathcal{T}_i$ is composed of two parts $[\mathcal{S}, \mathcal{Q}]$: (i) the support set $\mathcal{S}$, from which the model learns to adapt to the novel classes, and (ii) the query set $\mathcal{Q}$, on which the model is evaluated. The support set $\mathcal{S}=\{\bm{x}^{s}_i, y^{s}_i\}_{i=1}^{NK}$ is constructed by drawing $K$ labeled random samples from $N$ different classes, resulting in the so-called ($N$-way, $K$-shot) setting. The query set $\mathcal{Q}=\{\bm{x}^{q}_j\}_{j=1}^{NQ}$ contains $NQ$ (with $Q > K$) unlabeled samples. The base and novel classes are mutually exclusive, i.e., the distributions of $\mathcal{D}_{\textup{tr}}$ and $\mathcal{D}_{\textup{tst}}$ are different.

%

\section{Proposed Method: \ourmethod{}}
\label{sec:proposed-method}
\vspace{-0.3cm}

\subsection{Unsupervised Pretraining}
\label{ssec:self_sup}
\vspace{-0.2cm}
%
\begin{figure}[t]
\centering
\begin{overpic}[abs, unit=1cm, width=0.98\textwidth, clip, percent]{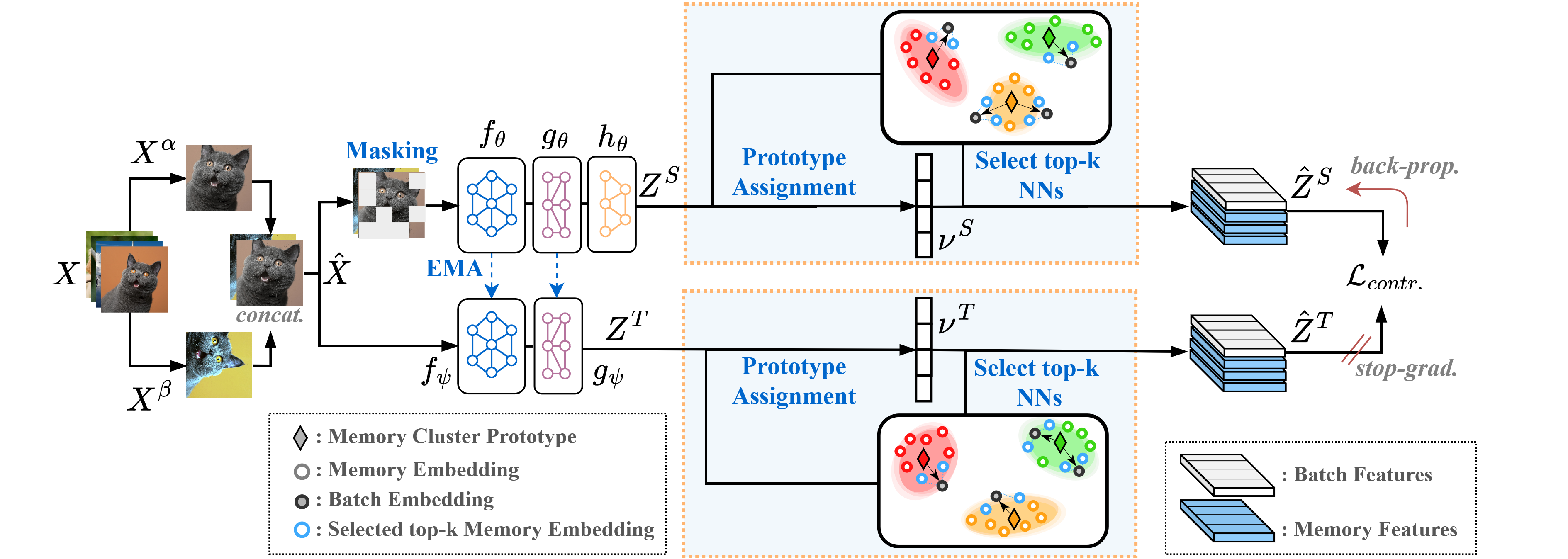}
    \put(45.3, 32.5){\normalsize \ourmodule{}$^S$}
    \put(45, 1){\normalsize \ourmodule{}$^T$}
    \end{overpic}
\caption{\small Overview of the proposed pretraining framework of \ourmethod{}. Two augmented views of the batch images $\bm{X}^{\{\alpha,\beta\}}$ are both passed through a student-teacher network followed by the \ourmodule{} memory module. \ourmodule{}  enhances the original batch with meaningful positives and dynamically updates the memory partitions.}
\label{fig:fig-pretrain}
\vspace{-0.5cm}
\end{figure}

\begin{wrapfigure}{r}{0.5\columnwidth}
    \centering
    \begin{overpic}[abs, unit=1cm, width=0.5\columnwidth, clip, percent]{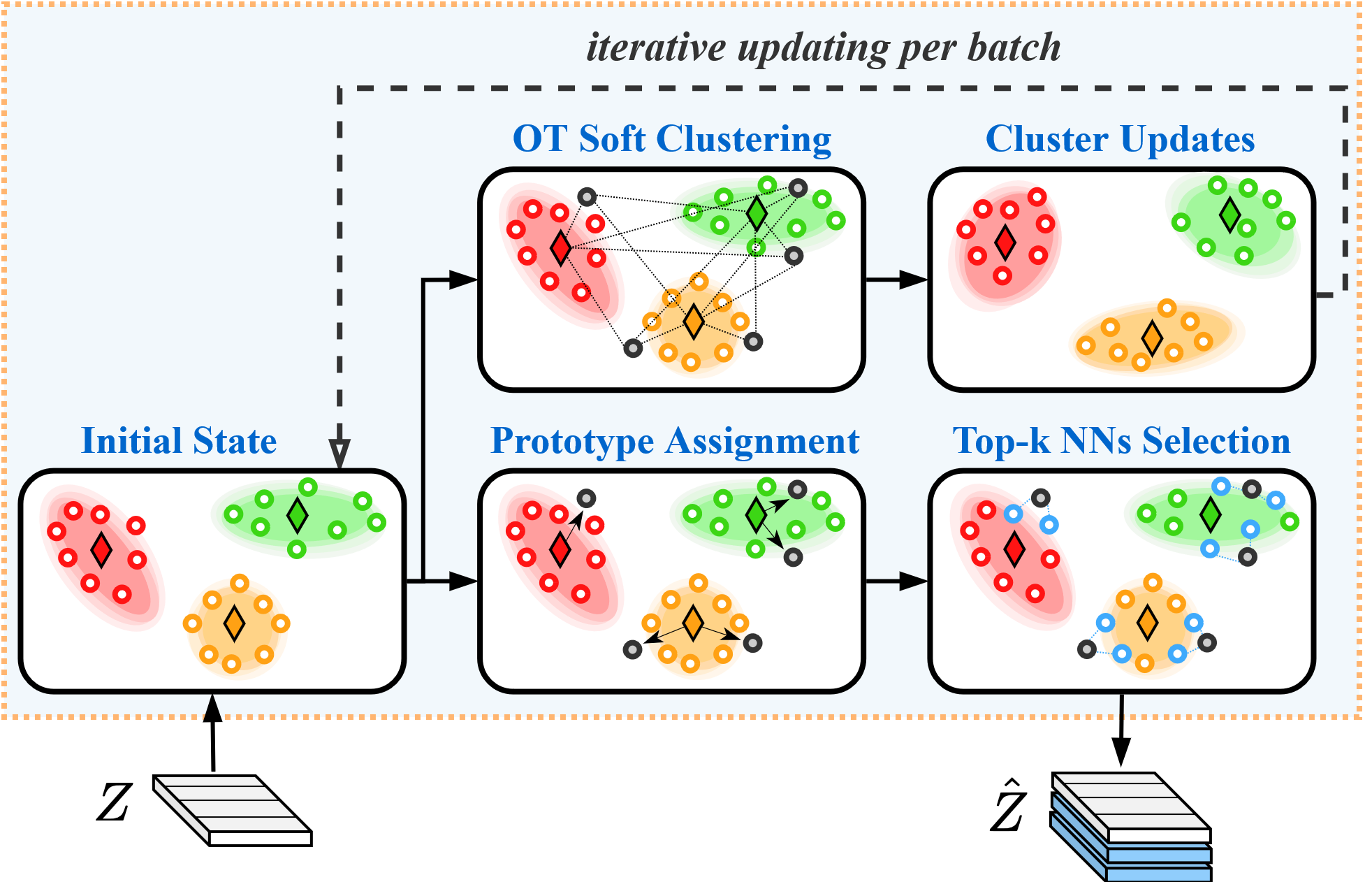}
        \put(2, 59){\normalsize \ourmodule{}}
        \end{overpic}
    \caption{\small Overview of the proposed dynamic clustered memory (\ourmodule{}) and its two informational paths.}
    \label{fig:dyce}
    \vspace{5pt}

    \begin{minipage}{0.5\textwidth}
        \LinesNumbered
        \IncMargin{1.2em} 
        \begin{algorithm}[H]
            \scriptsize
            \setstretch{1}
            \SetKwInput{Require}{Require}
            \SetKwInput{Return}{Return}
            \SetAlgoLined
        	\DontPrintSemicolon
        	\SetNoFillComment
            
            \Require{$\operatorname{epoch}_\textup{\texttt{thr}}$, $\mathcal{M}$, $\bm{\Gamma}$, $\bm{Z}$, $B$, $k$}
            
            \eIf{\:\:$|\mathcal{M}| = M$}{
            
                \textcolor{darkgreen}{/ / \:\:\textbf{Path (i)}: top-k and batch enhancement)}
                
                \If{\:\:$\operatorname{epoch}$ $\geq$ $\operatorname{epoch}_\textup{\texttt{thr}}$}{
                    
                     $\bm{\nu} = [\nu_i]_{i=1}^{2B}=\big[\underset{j \in \left[P\right]}{\texttt{argmin}} \langle\bm{z}_i, \bm{\gamma}_j\rangle\big]_{i=1}^{2B}$\; 
    
                     $\bm{Y}_i \gets \texttt{top-k}\big(\{\bm{z}_i, \mathcal{P}_{\nu_i}\}\big), \forall i \in \left[2B\right]$\;
                     
                     $\hat{\bm{Z}} = \left[\bm{Z},\bm{Y_1},\ldots,\bm{Y}_{2B}\right]$\; 
                }
                \textcolor{darkgreen}{/ / \:\:\textbf{Path (ii)}: iterative memory updating}
            
                \text{Find optimal plan between $\bm{Z}$, $\bm{\Gamma}$:} $\bm{\pi}^* \gets $ Solve Eq.~\ref{eqn:opt-transp}\;

                \text{Update $\mathcal{M}$ with new $\bm{Z}$:} \scalebox{0.95}{$\mathcal{M} \gets \texttt{update}(\mathcal{M},\bm{\pi}^*,\bm{Z})$}\; 
                
                \text{Discard $2B$ oldest batch embeddings:} $\texttt{dequeue}(\mathcal{M})$\;   
            }
            {
            \text{Store new batch:} $\mathcal{M} \gets \texttt{store}(\mathcal{M},\bm{Z})$\;
            }

            \Return {$\hat{\bm{Z}}$}
            \caption{\ourmodule{}}\label{alg:DyCE-ver1}
        \end{algorithm}
        \DecMargin{1.2em}
    \end{minipage}
    \vspace{-18pt}
\end{wrapfigure}

We build the unsupervised pretraining strategy of \ourmethod{} following contrastive representation learning. The core idea here is to efficiently attract ``positive'' samples (i.e., augmentations of the same image) in the representation space, while repelling ``negative'' samples. However, traditional contrastive learning approaches address this at the \emph{instance level}, where each image within the batch has to correspond to a unique class (a statistically unrealistic assumption!). As a result, potential positives present within a batch might be treated as negatives, which can have a detrimental impact on performance. A common strategy to combat this pitfall (also to avoid prohibitively large batch sizes) is to use a memory queue \citep{wu2018unsupervised, zhuang2019local, moco}. Exceptionally, 
PsCo \citep{psco} uses optimal transport to sample from a first-in-first-out memory queue and generate pseudo-supervised few-shot tasks in a meta-learning-based framework, whereas NNCLR \citep{nnclr} uses a nearest-neighbor approach to generate more meaningful positive pairs in the contrastive loss. However, these memory queues are still oblivious to global memberships (i.e., class-level information) in the latent space. Instead, we propose to infuse membership/class-level insights through a novel memory module (\ourmodule{}) within the pretraining phase of the proposed end-to-end approach: \ourmethod{}. Fig.~\ref{fig:fig-pretrain} provides a schematic illustration of the proposed contrastive pretraining framework within \ourmethod{}, and Fig.~\ref{fig:dyce} depicts the mechanics of \ourmodule{}.

\begin{wrapfigure}{r}{0.5\columnwidth}
    \vspace{-12pt}
     \begin{minipage}{0.5\textwidth}
            \LinesNumbered
            \IncMargin{1.2em} 
            \begin{algorithm}[H]
                \scriptsize
                \setstretch{1}
                \SetKwInput{Require}{Require}
                \SetKwInput{Return}{Return}
                \SetAlgoLined
            	\DontPrintSemicolon
            	\SetNoFillComment
             
                \Require{$\mathcal{A}$, $\theta$, $\psi$, $f_\theta$, $f_\psi$, $g_\theta$, $g_\psi$, $h_\theta$, $\mu$, $m$, \ourmodule{}$(\cdot)$}
                
                 $\hat{\bm{X}}=\big[\zeta^\alpha(\bm{X}),\zeta^\beta(\bm{X}) \big]$ for $\zeta^\alpha, \zeta^\beta \sim \mathcal{A}$\;
                
                $\bm{Z}^S = h_{\theta } \circ g_{\theta} \circ f_{\theta}\big(\mu(\hat{\bm{X}})\big)$\;

                $\bm{Z}^T = g_{\psi} \circ f_{\psi} (\hat{\bm{X}})$\;
                 
                 $\hat{\bm{Z}}^S, \; \hat{\bm{Z}}^T = \ourmodule{}^S(\bm{Z}^S), \; \ourmodule{}^T(\bm{Z}^T)$\;
                
                \text{Compute loss:} $\mathcal{L}_{contr.}$ using Eq.~\ref{eqn:finalloss} on $\hat{\bm{Z}}^S, \hat{\bm{Z}}^T$\; 

                \text{Update:} $\theta \gets \theta - \nabla\mathcal{L}_{contr.}, \; \psi \gets m\psi + (1-m)\theta$\; 
            
                
                \caption{Pretraining of \ourmethod{}}\label{alg:BECLR-ver1}
            \end{algorithm}
            \DecMargin{1.2em}
    \end{minipage}
\vspace{-12pt}
\end{wrapfigure}

\textbf{Pretraining Strategy of \ourmethod{}.} The pretraining pipeline of \ourmethod{} is summarized in Algorithm~\ref{alg:BECLR-ver1}, and a Pytorch-like pseudo-code can be found in Appendix~\ref{sec:appendix-pseudocode}. Let us now walk you through the algorithm. Let $\zeta^a,\zeta^b \sim \mathcal{A}$ be two randomly sampled data augmentations from the set of all available augmentations, $\mathcal{A}$. The mini-batch can then be denoted as $\hat{\bm{X}} = [\hat{\bm{x}}_i]_{i=1}^{2B} = \big[[\zeta^a(\bm{x}_i)]_{i=1}^B, [\zeta^b(\bm{x}_i)]_{i=1}^B\big]$, where $B$ the original batch size (line 1, Algorithm~\ref{alg:BECLR-ver1}). As shown in Fig.~\ref{fig:fig-pretrain}, we adopt a student-teacher (a.k.a Siamese) asymmetric momentum architecture similar to \citet{byol, simsiam}. Let $\mu(\cdot)$ be a patch-wise masking operator, $f(\cdot)$ the backbone feature extractor (ResNets \citep{resnets} in our case), and $g(\cdot)$, $h(\cdot)$ projection and prediction multi-layer perceptrons (MLPs), respectively. The teacher weight parameters ($\psi$) are an exponential moving averaged (\emph{EMA}) version of the student parameters ($\theta$), i.e., $\psi \gets m\psi + (1-m)\theta$, as in \citet{byol}, where $m$ the momentum hyperparameter, while $\theta$ are updated through stochastic gradient descent (SGD).
The student and teacher representations $\bm{Z}^S$ and $\bm{Z}^T$ (both of size $2B \times d$, with $d$ the latent embedding dimension) can then be obtained as follows:  $\bm{Z}^S = h_\theta \circ g_\theta \circ f_\theta \big(\mu (\hat{\bm{X}})\big), \:  \bm{Z}^T = g_\psi \circ f_\psi(\hat{\bm{X}})$  (lines 2,3).


Upon extracting $\bm{Z}^S$ and $\bm{Z}^T$, they are fed into the proposed dynamic memory module (\ourmodule{}), where enhanced versions of the batch representations $\hat{\bm{Z}}^S$, $\hat{\bm{Z}}^T$ (both of size $2B(k+1) \times d$, with $k$ denoting the number of selected nearest neighbors) are generated (line 4). Finally, we apply the contrastive loss in Eq.~\ref{eqn:finalloss} on the enhanced  batch representations $\bm{\hat{Z}^S}$, $\bm{\hat{Z}^T}$ (line 5). Upon finishing unsupervised pretraining, only the student encoder ($f_\theta$) is kept for the subsequent inference stage.

\textbf{Dynamic Clustered Memory (\ourmodule{}).} \emph{How do we manage to enhance the batch with meaningful true positives in the absence of labels?} We introduce \ourmodule{}: a novel dynamically updated clustered memory to moderate the representation space during training, while infusing a semblance of class-cognizance. We demonstrate later on in Section~\ref{sec:experimental-eval} that the design choices in \ourmodule{} have a significant impact on both pretraining performance as well as the downstream few-shot classification. 

Let us first establish some notation. We consider a memory unit $\mathcal{M}$ capable of storing up to $M$ latent embeddings (each of size $d$). To accommodate clustered memberships within \ourmodule{}, we consider up to $P$ partitions (or clusters) $\mathcal{P}_i$ in $\mathcal{M}=[\mathcal{P}_1,\ldots,\mathcal{P}_P]$, each of which is represented by a prototype stored in $\bm{\Gamma}=[\bm{\gamma}_1,\ldots,\bm{\gamma}_P]$. Each prototype $\bm{\gamma}_i$ (of size $1 \times d$) is the average of the latent embeddings stored in partition $\mathcal{P}_i$. As shown in Fig.~\ref{fig:dyce} and in Algorithm~\ref{alg:DyCE-ver1}, \ourmodule{} consists of two informational paths: (i) the top-$k$ neighbor selection and batch enhancement path (bottom branch of the figure), which uses the current state of $\mathcal{M}$ and $\bm{\Gamma}$; (ii) the iterative memory updating via dynamic clustering path (top branch). \ourmodule{} takes as input student or teacher embeddings (we use $\bm{Z}$ here, for brevity) and returns the enhanced versions $\hat{\bm{Z}}$. We also allow for an adaptation period $\operatorname{epoch} < \operatorname{epoch}_\textup{\texttt{thr}}$ (empirically $20$-$50$ epochs), during which path (i) is not activated and the training batch is not enhanced. To further elaborate, path (i) starts with assigning each $\bm{z}_i \in \bm{Z}$ to its nearest (out of $P$) memory prototype $\bm{\gamma}_{\nu_i}$ based on the Euclidean distance \scalebox{0.95}{$\left\langle \cdot \right\rangle$}. $\bm{\nu}$ is a vector of indices (of size $2B \times 1$) that contains these prototype assignments for all batch embeddings (line 4, Algorithm~\ref{alg:DyCE-ver1}). Next (in line 5), we select  the $k$ most similar memory embeddings to $\bm{z}_i$ from the memory partition corresponding to its assigned prototype ($\mathcal{P}_{\nu_i}$) and store them in $\bm{Y}_i$ (of size $k \times d$). Finally (in line 6), all $\bm{Y}_i, \forall i \in [2B]$ are concatenated into the enhanced batch $\hat{\bm{Z}} = [\bm{Z},\bm{Y_1},\ldots,\bm{Y}_{2B}]$ of size $L \times d$ (where $L=2B(k+1)$). Path (ii) addresses the iterative memory updating by casting it into an optimal transport problem \citep{sinkhorn} given by: 
\begin{equation} \label{eqn:transport}
\small
\bm{\Pi}(\bm{r}, \bm{c})=\left\{\,\bm{\pi} \in \mathbb{R}_{+}^{2B \: \times \: P} \mid \bm{\pi} \mathbf{1}_{P}=\bm{r}, \: \bm{\pi}^{\top} \mathbf{1}_{2B}=\bm{c}, \; \bm{r} = \mathbf{1} \cdot \sfrac{1}{2B}, \;\bm{c} = \mathbf{1} \cdot \sfrac{1}{P}\,\right\}, 
\end{equation}
to find a transport plan $\bm{\pi}$ (out of $\bm{\Pi}$) mapping $\bm{Z}$ to $\bm{\Gamma}$. Here, $\bm{r} \in \mathbb{R}^{2B}$ denotes the distribution of batch embeddings $[\bm{z}_i]_{i=1}^{2B}$, $\bm{c} \in \mathbb{R}^{P}$ is the distribution of memory cluster prototypes $[\bm{\gamma}_{i}]_{i=1}^{P}$. The last two conditions in Eq.~\ref{eqn:transport} enforce equipartitioning (i.e., uniform assignment) of $\bm{Z}$ into the $P$ memory partitions/clusters. Obtaining the optimal transport plan, $\bm{\pi}^\star$, can then be formulated as:
\begin{equation}\label{eqn:opt-transp}
\small
\bm{\pi}^{\star}=\underset{\bm{\pi} \in \bm{\Pi}\left(\bm{r}, \bm{c} \right)}{\operatorname{argmin}}\left\langle\bm{\pi}, \bm{D}\right\rangle_{F} - \varepsilon \mathbb{H}(\bm{\pi}),
\end{equation}
and solved using the Sinkhorn-Knopp \citep{sinkhorn} algorithm (line 8). Here, $\bm{D}$ is a pairwise distance matrix between the elements of ${\bm Z}$ and ${\bm \Gamma}$ (of size $2B \times P$), \scalebox{0.95}{$\left\langle \cdot \right\rangle_F$} denotes the Frobenius dot product, $\varepsilon$ is a regularisation term, and $\mathbb{H}(\cdot)$ is the Shannon entropy. Next (in line 9), we add the embeddings of the current batch $\bm{Z}$ to $\mathcal{M}$ and use $\bm{\pi}^{\star}$ for updating the partitions $\mathcal{P}_i$ and prototypes $\bm{\Gamma}$ (using EMA for updating). Finally, we discard the $2B$ oldest memory embeddings (line 10). 

\textbf{Loss Function.} The popular infoNCE loss \citep{infonce} is the basis of our loss function, yet recent studies \citep{bias_ince_1,bias_ince_2} have shown that it is prone to high bias, when the batch size is small. To address this, we adopt a variant of infoNCE, which maximizes the same mutual information objective, but has been shown to be less biased \citep{unisiam}:
\begin{equation}\label{eqn:finalloss}
\small
\mathcal{L}_{contr.} = \frac{1}{L} \sum_{i=1}^{\sfrac{L}{2}} \bigg(\texttt{d}[\bm{z}^S_{i}, \bm{z}^{T+}_{i}] + \texttt{d}[\bm{z}^{S+}_{i}, \bm{z}^T_{i}]\bigg)  - \: \lambda   \log \bigg(\frac{1}{L}\sum_{i=1}^{L} \sum_{j \neq i,i^+}\exp (\texttt{d}[\bm{z}^S_{i}, \bm{z}^S_{j}]/ \tau)\bigg),
\end{equation}
where $\tau$ is a temperature parameter,  $\texttt{d}$$[\cdot]$ is the negative cosine similarity, $\lambda$ is a weighting hyperparameter, $L$ is the enhanced batch size and $\bm{z}_i^{+}$ stands for the latent embedding of the positive sample, corresponding to sample $i$. Following \citep{simsiam}, to further boost training performance, we pass both views through both the student and the teacher. The first term in Eq.~\ref{eqn:finalloss} operates on positive pairs, and the second term pushes each representation away from all other batch representations.

\subsection{Supervised Inference}
\label{ssec:sup_inference}
\vspace{-0.2cm}
Supervised inference (a.k.a fine-tuning) usually combats the distribution shift between training and test datasets. However, the limited number of support samples (in FSL tasks) at test time leads to a significant performance degradation due to the so-called \emph{sample bias} \citep{cui2021parameterless,xu2022alleviating}. This issue is mostly disregarded in recent state-of-the-art U-FSL baselines \citep{pdanet,unisiam,metadm_uni}. As part of the inference phase of \ourmethod{}, we propose a simple, yet efficient, add-on module (coined as \ourft{}) for aligning the distributions of the query ($\mathcal{Q}$) and support ($\mathcal{S}$) sets, to structurally address sample bias.  Notice that \ourft{} is not a learnable module and that there are no model updates nor any fine-tuning involved in the inference stage of \ourmethod{}.
\begin{figure}[t]
\centering
    \begin{overpic}[abs, unit=1cm, width=0.95\textwidth, clip, percent]{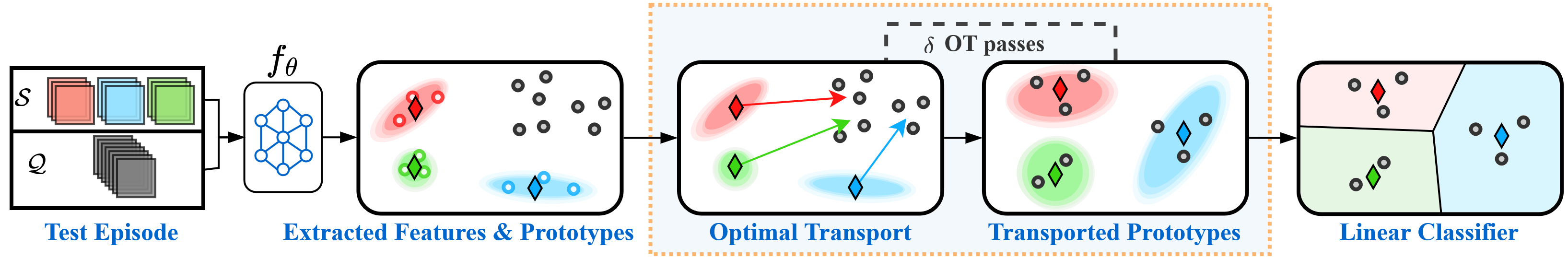}
        \put(42, 14){\small \ourft{}}
    \end{overpic}
 \caption{\small Overview of the inference strategy of \ourmethod{}. Given a test episode, the support ($\mathcal{S}$) and query ($\mathcal{Q}$) sets are passed to the pretrained feature extractor ($f_\theta$). \ourft{} aligns support prototypes and query features.}
 \label{fig:fine-tuning}
 \vspace{-0.2cm}
\end{figure}

\textbf{Optimal Transport-based Distribution Alignment (\ourft{}).}\label{sssec:opt-aln} 
Let $\mathcal{T} = \mathcal{S} \cup \mathcal{Q}$ be a downstream few-shot task. We first extract the support $\bm{Z}^\mathcal{S}=f_{\theta}(\mathcal{S})$ (of size $NK \times d$) and query $\bm{Z}^\mathcal{Q}=f_{\theta}(\mathcal{Q})$ (of size $NQ \times d$) embeddings and calculate the support set prototypes $\bm{P}^{\mathcal{S}}$ (class averages of size $N \times d$). Next, we find the optimal transport plan ($\bm{\pi}^{\star}$) between $\bm{P}^{\mathcal{S}}$ 
and $\bm{Z}^\mathcal{Q}$ using Eq.~\ref{eqn:opt-transp}, with $\bm{r} \in \mathbb{R}^{NQ}$ the distribution of $\bm{Z}^\mathcal{Q}$ and $\bm{c} \in \mathbb{R}^{N}$ the distribution of $\bm{P}^{\mathcal{S}}$. Finally, we use $\bm{\pi}^{\star}$ to map the support set prototypes onto the region occupied by the query embeddings:
\begin{equation}
\begin{aligned}\label{eqn:bcenter-map}
\scriptsize
\hat{\bm{P}}^{\mathcal{S}} &= \hat{\bm{\pi}}^{\star T} \bm{Z}^\mathcal{Q}, \:\:\: \hat{\bm{\pi}}^{\star}_{i, j} = \frac{\bm{\pi}^{\star}_{i, j}}{\sum_{j} {\bm{\pi}}^{\star}_{i, j}}, \forall i \in [NQ], j \in [N], 
\end{aligned}
\end{equation}
where $\hat{\bm{\pi}}^{\star}$ is the normalized transport plan and $\hat{\bm{P}}^{\mathcal{S}}$ are the transported support prototypes. Finally, we fit a logistic regression classifier on $\hat{\bm{P}}^{\mathcal{S}}$ to infer on the unlabeled query set. We show in Section~\ref{sec:experimental-eval} that \ourft{} successfully minimizes the distribution shift (between support and query sets) and contributes to the overall significant performance margin \ourmethod{} offers against the best existing baselines. Note that we iteratively perform $\delta$ consecutive passes of \ourft{}, where $\hat{\bm{P}}^{\mathcal{S}}$ acts as the input of the next pass. Notably, \ourft{} can straightforwardly be applied on top of any U-FSL approach. An overview of \ourft{} and the proposed inference strategy is illustrated in Fig.~\ref{fig:fine-tuning}.

\textbf{Remark:} \ourft{} operates on two distributions and relies on the total number of unlabeled query samples being larger than the total number of labeled support samples ($|\mathcal{Q}| > |\mathcal{S}|$) for reasonable distribution mapping, which is also the standard convention in the U-FSL literature. That said, \ourft{} would still perform on imbalanced FSL tasks as long as the aforementioned condition is met.



\section{Experiments}
\label{sec:experimental-eval}
\vspace{-0.2cm}
In this section, we rigorously study the performance of the proposed approach both quantitatively as well as qualitatively by addressing the following three main questions:\\ $[$\textbf{Q1}$]$ How does \ourmethod{} perform against the state-of-the-art for \emph{in-domain} and \emph{cross-domain} settings?\\ $[$\textbf{Q2}$]$ Does \ourmodule{} affect pretraining performance by establishing \emph{separable memory partitions}?\\ $[$\textbf{Q3}$]$ Does \ourft{} address the \emph{sample bias} via the proposed distribution alignment strategy?

We use PyTorch \citep{paszke2019pytorch} for all implementations. Elaborate implementation and training details are discussed in the supplementary material, in Appendix~\ref{sec:appendix-impl-det}.

\textbf{Benchmark Datasets.} We evaluate \ourmethod{} in terms of its in-domain performance on the two most widely adopted few-shot image classification datasets: miniImageNet \citep{miniImageNet} and tieredImageNet \citep{tieredImageNet}. Additionally, for the in-domain setting, we also evaluate on two curated versions of CIFAR-$100$ \citep{cifar} for FSL, i.e., CIFAR-FS and FC$100$. Next, we evaluate \ourmethod{} in cross-domain settings on the Caltech-UCSD Birds (CUB) dataset \citep{cub} and a more recent cross-domain FSL (CDFSL) benchmark \citep{cdfsl}. For cross-domain experiments, miniImageNet is used as the pretraining (source) dataset and ChestX \citep{chestx}, ISIC \citep{isic}, EuroSAT \citep{eurosat} and CropDiseases \citep{cropdisease} (in Table~\ref{tab:cdfsl}) and CUB (in Table~\ref{tab:mini to CUB-full} in the Appendix), as the inference (target) datasets. 

\begin{table}[t]
\centering
\aboverulesep = 0pt
\belowrulesep = 0pt
\renewcommand{\arraystretch}{1.2}
\caption{\small Accuracies (in \% $\pm$ std) on miniImageNet and tieredImageNet compared against unsupervised (\texttt{Unsup.}) and supervised (\texttt{Sup.}) baselines. Backbones: \texttt{RN}: Residual network.$^\dagger$: denotes our reproduction. $^*$: denotes extra synthetic training data used. Style: \textbf{best} and \underline{second best}.}
\vspace{-0.1cm}
\label{tab:mini-tiered}
\resizebox{\textwidth}{!}{%
\begin{tabular}{@{}lcccccc@{}}
\toprule
 &
   &
  \multicolumn{1}{c|}{} &
  \multicolumn{2}{c|}{\cellcolor[HTML]{DCE7D9}\textbf{miniImageNet}} &
  \multicolumn{2}{c}{\cellcolor[HTML]{FDE5C9}\textbf{tieredImageNet}} \\ \midrule
\multicolumn{1}{l|}{\textbf{Method}} &
  \multicolumn{1}{c|}{\ \  \ \textbf{Backbone}\ \  \ } &
  \multicolumn{1}{c|}{\textbf{Setting}} &
  \multicolumn{1}{c|}{\ \  \ \textbf{5-way 1-shot}\ \ \ } &
  \multicolumn{1}{c|}{\ \ \ \textbf{5-way 5-shot}\ \ \ } &
  \multicolumn{1}{c|}{\ \ \ \textbf{5-way 1-shot}\ \ \ } &
  \ \ \ \textbf{5-way 5-shot}\ \ \  \\ \midrule
  \multicolumn{1}{l|}{SwAV$^\dagger$ \citep{SwAV}} &
  \multicolumn{1}{c|}{$\texttt{RN18}$} &
  \multicolumn{1}{c|}{$\texttt{Unsup.}$} &
  \multicolumn{1}{c|}{59.84 \scriptsize{± 0.52}} &
  \multicolumn{1}{c|}{78.23 \scriptsize{± 0.26}} &
  \multicolumn{1}{c|}{65.26 \scriptsize{± 0.53}} &
  81.73 \scriptsize{± 0.24} \\
\multicolumn{1}{l|}{NNCLR$^\dagger$ \citep{nnclr}} &
  \multicolumn{1}{c|}{$\texttt{RN18}$} &
  \multicolumn{1}{c|}{$\texttt{Unsup.}$} &
  \multicolumn{1}{c|}{63.33 \scriptsize{± 0.53}} &
  \multicolumn{1}{c|}{80.75 \scriptsize{± 0.25}} &
  \multicolumn{1}{c|}{65.46 \scriptsize{± 0.55}} &
  81.40 \scriptsize{± 0.27} \\
\multicolumn{1}{l|}{CPNWCP \citep{wang2022contrastive}} &
  \multicolumn{1}{c|}{$\texttt{RN18}$} &
  \multicolumn{1}{c|}{$\texttt{Unsup.}$} &
  \multicolumn{1}{c|}{53.14 \scriptsize{± 0.62}} &
  \multicolumn{1}{c|}{67.36 \scriptsize{± 0.5}} &
  \multicolumn{1}{c|}{45.00 \scriptsize{± 0.19}} &
  62.96 \scriptsize{± 0.19} \\
\multicolumn{1}{l|}{HMS \citep{hms}} &
  \multicolumn{1}{c|}{$\texttt{RN18}$} &
  \multicolumn{1}{c|}{$\texttt{Unsup.}$} &
  \multicolumn{1}{c|}{58.20 \scriptsize{± 0.23}} &
  \multicolumn{1}{c|}{75.77 \scriptsize{± 0.16}} &
  \multicolumn{1}{c|}{58.42 \scriptsize{± 0.25}} &
  75.85 \scriptsize{± 0.18} \\
\multicolumn{1}{l|}{\textcolor{black}{SAMPTransfer$^\dagger$} \citep{samptransfer}} &
  \multicolumn{1}{c|}{\textcolor{black}{$\texttt{RN18}$}} &
  \multicolumn{1}{c|}{\textcolor{black}{$\texttt{Unsup.}$}} &
    \multicolumn{1}{c|}{\textcolor{black}{45.75 \scriptsize{± 0.77}}} &
  \multicolumn{1}{c|}{\textcolor{black}{68.33 \scriptsize{± 0.66}}} &
  \multicolumn{1}{c|}{\textcolor{black}{42.32 \scriptsize{± 0.75}}} &
  \textcolor{black}{53.45 \scriptsize{± 0.76}} \\
\multicolumn{1}{l|}{\textcolor{black}{PsCo}$^\dagger$ \citep{psco}} &
  \multicolumn{1}{c|}{\textcolor{black}{$\texttt{RN18}$}} &
  \multicolumn{1}{c|}{\textcolor{black}{$\texttt{Unsup.}$}} &
  \multicolumn{1}{c|}{\textcolor{black}{47.24 \scriptsize{± 0.76}}} &
  \multicolumn{1}{c|}{\textcolor{black}{65.48 \scriptsize{± 0.68}}} &
  \multicolumn{1}{c|}{\textcolor{black}{54.33 \scriptsize{± 0.54}}} &
  \textcolor{black}{69.73 \scriptsize{± 0.49}} \\
\multicolumn{1}{l|}{UniSiam + dist \citep{unisiam}} &
  \multicolumn{1}{c|}{$\texttt{RN18}$} &
  \multicolumn{1}{c|}{$\texttt{Unsup.}$} &
  \multicolumn{1}{c|}{64.10 \scriptsize{± 0.36}} &
  \multicolumn{1}{c|}{82.26 \scriptsize{± 0.25}} &
  \multicolumn{1}{c|}{67.01 \scriptsize{± 0.39}} &
  {\ul 84.47} \scriptsize{± 0.28} \\
\multicolumn{1}{l|}{Meta-DM + UniSiam + dist$^*$ \citep{metadm_uni} \ \  \ \ \ \ \ \ \ \ \ \ \ } &
  \multicolumn{1}{c|}{ \ \ \ \ $\texttt{RN18}$ \ \ \ \ } &
  \multicolumn{1}{c|}{ \ \ \ \ $\texttt{Unsup.}$ \ \ \ \ } &
  \multicolumn{1}{c|}{{\ul 65.64} \scriptsize{± 0.36}} &
  \multicolumn{1}{c|}{{\ul 83.97} \scriptsize{± 0.25}} &
  \multicolumn{1}{c|}{{\ul 67.11} \scriptsize{± 0.40}} &
  84.39 \scriptsize{± 0.28} \\ [1pt] \cdashlinelr{1-7} 
  \multicolumn{1}{l|}{MetaOptNet \citep{metaoptnet}} &
  \multicolumn{1}{c|}{$\texttt{RN18}$} &
  \multicolumn{1}{c|}{$\texttt{Sup.}$} &
  \multicolumn{1}{c|}{64.09 \scriptsize{± 0.62}} &
  \multicolumn{1}{c|}{80.00 \scriptsize{± 0.45}} &
  \multicolumn{1}{c|}{65.99 \scriptsize{± 0.72}} &
  81.56 \scriptsize{± 0.53} \\
\multicolumn{1}{l|}{Transductive CNAPS \citep{transdcnaps}} &
  \multicolumn{1}{c|}{$\texttt{RN18}$} &
  \multicolumn{1}{c|}{$\texttt{Sup.}$} &
  \multicolumn{1}{c|}{55.60 \scriptsize{± 0.90}} &
  \multicolumn{1}{c|}{73.10 \scriptsize{± 0.70}} &
  \multicolumn{1}{c|}{65.90 \scriptsize{± 1.10}} &
  81.80 \scriptsize{± 0.70} \\
\midrule
\rowcolor[HTML]{E7F5F8} 
\multicolumn{1}{l|}{\cellcolor[HTML]{E7F5F8}\textbf{\ourmethod{} \textbf{(Ours)}}} &
  \multicolumn{1}{c|}{\cellcolor[HTML]{E7F5F8}$\texttt{RN18}$} &
  \multicolumn{1}{c|}{\cellcolor[HTML]{E7F5F8}$\texttt{Unsup.}$} &
  \multicolumn{1}{c|}{\cellcolor[HTML]{E7F5F8}\textbf{75.74} \scriptsize{± 0.62}} &
  \multicolumn{1}{c|}{\cellcolor[HTML]{E7F5F8}\textbf{84.93} \scriptsize{± 0.33}} &
  \multicolumn{1}{c|}{\cellcolor[HTML]{E7F5F8}\textbf{76.44} \scriptsize{± 0.66}} &
  \multicolumn{1}{c}{\cellcolor[HTML]{E7F5F8}\textbf{84.85} \scriptsize{± 0.37}} \\ \bottomrule
\multicolumn{1}{l|}{PDA-Net \citep{pdanet}} &
  \multicolumn{1}{c|}{$\texttt{RN50}$} &
  \multicolumn{1}{c|}{$\texttt{Unsup.}$} &
  \multicolumn{1}{c|}{63.84 \scriptsize{± 0.91}} &
  \multicolumn{1}{c|}{83.11 \scriptsize{± 0.56}} &
  \multicolumn{1}{c|}{69.01 \scriptsize{± 0.93}} &
  84.20 \scriptsize{± 0.69} \\
\multicolumn{1}{l|}{UniSiam + dist \citep{unisiam}} &
  \multicolumn{1}{c|}{$\texttt{RN50}$} &
  \multicolumn{1}{c|}{$\texttt{Unsup.}$} &
  \multicolumn{1}{c|}{65.33 \scriptsize{± 0.36}} &
  \multicolumn{1}{c|}{83.22 \scriptsize{± 0.24}} &
  \multicolumn{1}{c|}{69.60 \scriptsize{± 0.38}} &
  86.51 \scriptsize{± 0.26} \\
\multicolumn{1}{l|}{Meta-DM + UniSiam + dist$^*$ \citep{metadm_uni}} &
  \multicolumn{1}{c|}{$\texttt{RN50}$} &
  \multicolumn{1}{c|}{$\texttt{Unsup.}$} &
  \multicolumn{1}{c|}{{\ul 66.68} \scriptsize{± 0.36}} &
  \multicolumn{1}{c|}{{\ul 85.29} \scriptsize{± 0.23}} &
  \multicolumn{1}{c|}{{\ul 69.61} \scriptsize{± 0.38}} &
  {\ul 86.53} \scriptsize{± 0.26} \\ \midrule
  [1pt] \cdashlinelr{1-7} 
\rowcolor[HTML]{E7F5F8} 
\multicolumn{1}{l|}{\cellcolor[HTML]{E7F5F8}\textbf{\ourmethod{} \textbf{(Ours)}}} &
  \multicolumn{1}{c|}{\cellcolor[HTML]{E7F5F8}$\texttt{RN50}$} &
  \multicolumn{1}{c|}{\cellcolor[HTML]{E7F5F8}$\texttt{Unsup.}$} &
  \multicolumn{1}{c|}{\cellcolor[HTML]{E7F5F8}\textbf{80.57} \scriptsize{± 0.57}} &
  \multicolumn{1}{c|}{\cellcolor[HTML]{E7F5F8}\textbf{87.82} \scriptsize{± 0.29}} &
  \multicolumn{1}{c|}{\cellcolor[HTML]{E7F5F8}\textbf{81.69} \scriptsize{± 0.61}} &
  \multicolumn{1}{c}{\cellcolor[HTML]{E7F5F8}\textbf{87.86} \scriptsize{± 0.32}} \\ \bottomrule
\end{tabular}%
}
\vspace{-3mm}
\end{table}
%

\subsection{Evaluation Results}
\label{ssec:eval-results}
\vspace{-0.2cm}
We report test accuracies with $95\%$ confidence intervals over $2000$ test episodes, each with $Q = 15$ query shots per class, for all datasets, as is most commonly adopted in the literature \citep{pdanet,deepeigenmaps,unisiam}. The performance on miniImageNet, tieredImageNet, CIFAR-FS, FC$100$ and miniImageNet $\rightarrow$ CUB is evaluated on ($5$-way, $\{1,5\}$-shot) classification tasks, whereas for miniImageNet $\rightarrow$ CDFSL we test on ($5$-way, $\{5,20\}$-shot) tasks, as is customary across the literature \citep{cdfsl, ericsson2021well}. We assess \ourmethod{}'s performance against a wide variety of baselines ranging from (i) established SSL baselines \citep{simclr,byol,SwAV,barlowtwins,simsiam,nnclr} to (ii) state-of-the-art U-FSL approaches \citep{pdanet,unisiam,samptransfer,deepeigenmaps,metadm_uni,psco}, as well as (iii) against a set of competitive supervised baselines \citep{LEO,ccrot,metaoptnet,transdcnaps}.

\begin{wraptable}{r}{0.55\columnwidth}
\vspace{-12pt}
\centering
\aboverulesep = 0pt
\belowrulesep = 0pt
\renewcommand{\arraystretch}{1.2}
\caption{\small Accuracies in (\% $\pm$ std) on CIFAR-FS and FC$100$ in ($5$-way, $\{1,5\}$-shot). Style: \textbf{best} and \underline{second best}.}
\vspace{-0.1cm}
\label{tab:cifar-main}
\resizebox{0.55\columnwidth}{!}{%
\begin{tabular}{@{}l|cccccc@{}}
\toprule
\textbf{} &
  \multicolumn{2}{c|}{\cellcolor[HTML]{DCE7D9}\textbf{CIFAR-FS}} &
  \multicolumn{2}{c}{\cellcolor[HTML]{FDE5C9}\textbf{FC100}} \\ \midrule
\multicolumn{1}{l|}{\textbf{Method}} &
  \multicolumn{1}{c|}{\textbf{1-shot}} &
  \multicolumn{1}{c|}{\textbf{5-shot}} &
  \multicolumn{1}{c|}{\textbf{1-shot}} &
  \textbf{5-shot} \\ \midrule  
\multicolumn{1}{l|}{SimCLR \citep{simclr}} &
  \multicolumn{1}{c|}{54.56 \tiny{± 0.19}} &
  \multicolumn{1}{c|}{71.19 \tiny{± 0.18}} &
  \multicolumn{1}{c|}{36.20 \tiny{± 0.70}} &
  49.90 \tiny{± 0.70} \\ 
\multicolumn{1}{l|}{MoCo v2 \citep{mocov2}} &
  \multicolumn{1}{c|}{52.73 \tiny{± 0.20}} &
  \multicolumn{1}{c|}{67.81 \tiny{± 0.19}} &
  \multicolumn{1}{c|}{37.70 \tiny{± 0.70}} &
  53.20 \tiny{± 0.70} \\ 
\multicolumn{1}{l|}{LF2CS \citep{lf2cs}} &
  \multicolumn{1}{c|}{{\ul 55.04} \tiny{± 0.72}} &
  \multicolumn{1}{c|}{70.62 \tiny{± 0.57}} &
  \multicolumn{1}{c|}{37.20 \tiny{± 0.70}} &
  52.80 \tiny{± 0.60} \\
\multicolumn{1}{l|}{Barlow Twins \citep{barlowtwins}} &
  \multicolumn{1}{c|}{-} &
  \multicolumn{1}{c|}{-} &
  \multicolumn{1}{c|}{37.90 \tiny{± 0.70}} &
  54.10 \tiny{± 0.60} \\ 
\multicolumn{1}{l|}{HMS \citep{hms}} &
  \multicolumn{1}{c|}{54.65 \tiny{± 0.20}} &
  \multicolumn{1}{c|}{{\ul 73.70} \tiny{± 0.18}} &
  \multicolumn{1}{c|}{37.88 \tiny{± 0.16}} &
  53.68 \tiny{± 0.18} \\ 
\multicolumn{1}{l|}{Deep Eigenmaps \citep{deepeigenmaps}} &
  \multicolumn{1}{c|}{-} &
  \multicolumn{1}{c|}{-} &
  \multicolumn{1}{c|}{{\ul 39.70} \tiny{± 0.70}} &
  {\ul 57.90} \tiny{± 0.70} \\ \midrule
\multicolumn{1}{l|}{\cellcolor[HTML]{E7F5F8}\textbf{\ourmethod{} \textbf{(Ours)}}} &
  \multicolumn{1}{c|}{\cellcolor[HTML]{E7F5F8}\textbf{70.39} \tiny{± 0.62}} &
  \multicolumn{1}{c|}{\cellcolor[HTML]{E7F5F8}\textbf{81.56} \tiny{± 0.39}} &
  \multicolumn{1}{c|}{\cellcolor[HTML]{E7F5F8}\textbf{45.21} \tiny{± 0.50}} &
  \multicolumn{1}{c}{\cellcolor[HTML]{E7F5F8}\textbf{60.02} \tiny{± 0.43}} \\ \bottomrule
\end{tabular}%
}
\vspace{-12pt}
\end{wraptable}
\textbf{$[$A1-a$]$ In-Domain Setting.}\label{sssec:in_domain} The results for miniImageNet and tieredImageNet in the ($5$-way, $\{1,5\}$-shot) settings are reported in Table~\ref{tab:mini-tiered}. Regardless of backbone depth, \ourmethod{} sets a \emph{new state-of-the-art} on both datasets, showing up to a $14\%$ and $2.5\%$ gain on miniImageNet over the prior art of U-FSL for the $1$ and $5$-shot settings, respectively. The results on tieredImageNet also highlight a considerable performance margin. Interestingly, \ourmethod{} even outperforms U-FSL baselines trained with extra (synthetic) training data, sometimes distilled from deeper backbones, also the cited supervised baselines. Table~\ref{tab:cifar-main} provides further insights on (the less commonly adopted) CIFAR-FS and FC$100$ benchmarks, showing a similar trend with up to $15\%$ and $8\%$ in $1$ and $5$-shot settings, respectively, for CIFAR-FS, and $5.5\%$ and $2\%$ for FC$100$.  

\begin{table}[t]
\centering
\aboverulesep = 0pt
\belowrulesep = 0pt
\renewcommand{\arraystretch}{1.2}
\caption{\small Accuracies (in \% $\pm$ std) on miniImageNet → CDFSL. $^\dagger$: our reproduc. Style: \textbf{best} and \underline{second best}.}
\vspace{-0.1cm}
\label{tab:cdfsl}
\resizebox{\textwidth}{!}{%
\begin{tabular}{@{}l|cc|cc|cc|cc@{}}
\toprule
\textbf{} &
  \multicolumn{2}{c|}{\cellcolor[HTML]{DCE7D9}\textbf{ChestX}} &
  \multicolumn{2}{c|}{\cellcolor[HTML]{FDE5C9}\textbf{ISIC}} &
  \multicolumn{2}{c|}{\cellcolor[HTML]{ECF4FF}\textbf{EuroSAT}} &
  \multicolumn{2}{c}{\cellcolor[HTML]{dedee0}\textbf{CropDiseases}} \\ \midrule
\multicolumn{1}{l|}{\textbf{Method}} &
  \multicolumn{1}{c|}{\textbf{5 way 5-shot}} &
  \textbf{5 way 20-shot} &
  \multicolumn{1}{c|}{\textbf{5 way 5-shot}} &
  \textbf{5 way 20-shot} &
  \multicolumn{1}{c|}{\textbf{5 way 5-shot}} &
  \textbf{5 way 20-shot} &
  \multicolumn{1}{c|}{\textbf{5 way 5-shot}} &
  \textbf{5 way 20-shot} \\ \midrule
\multicolumn{1}{l|}{SwAV$^\dagger$ \citep{SwAV}} &
  \multicolumn{1}{c|}{25.70 \scriptsize{± 0.28}} &
  30.41 \scriptsize{± 0.25} &
  \multicolumn{1}{c|}{40.69 \scriptsize{± 0.34}} &
  49.03 \scriptsize{± 0.30} &
  \multicolumn{1}{c|}{84.82 \scriptsize{± 0.24}} &
  90.77 \scriptsize{± 0.26} &
  \multicolumn{1}{c|}{ 88.64 \scriptsize{± 0.26}} &
  95.11 \scriptsize{± 0.21} \\
\multicolumn{1}{l|}{NNCLR$^\dagger$ \citep{nnclr}} &
  \multicolumn{1}{c|}{25.74 \scriptsize{± 0.41}} &
  29.54 \scriptsize{± 0.45} &
  \multicolumn{1}{c|}{38.85 \scriptsize{± 0.56}} &
  47.82 \scriptsize{± 0.53} &
  \multicolumn{1}{c|}{83.45 \scriptsize{± 0.57}} &
  90.80 \scriptsize{± 0.39} &
  \multicolumn{1}{c|}{ 90.76 \scriptsize{± 0.57}} &
  95.37 \scriptsize{± 0.37} \\
\multicolumn{1}{l|}{SAMPTransfer \citep{samptransfer}} &
  \multicolumn{1}{c|}{26.27 \scriptsize{± 0.44}} &
  \textbf{34.15} \scriptsize{± 0.50} &
  \multicolumn{1}{c|}{{\ul 47.60} \scriptsize{± 0.59}} &
  \textbf{61.28} \scriptsize{± 0.56} &
  \multicolumn{1}{c|}{85.55 \scriptsize{± 0.60}} &
  88.52 \scriptsize{± 0.50} &
  \multicolumn{1}{c|}{91.74 \scriptsize{± 0.55}} &
  96.36 \scriptsize{± 0.28} \\
  \multicolumn{1}{l|}{PsCo \citep{psco}} &
  \multicolumn{1}{c|}{24.78 \scriptsize{± 0.23}} &
  27.69 \scriptsize{± 0.23} &
  \multicolumn{1}{c|}{44.00 \scriptsize{± 0.30}} &
  54.59 \scriptsize{± 0.29} &
  \multicolumn{1}{c|}{81.08 \scriptsize{± 0.35}} &
  87.65 \scriptsize{± 0.28} &
  \multicolumn{1}{c|}{88.24 \scriptsize{± 0.31}} &
  94.95 \scriptsize{± 0.18} \\
\multicolumn{1}{l|}{UniSiam + dist \citep{unisiam}} &
  \multicolumn{1}{c|}{ \textbf{ 28.18} \scriptsize{± 0.45}} &
  \textbf{34.58} \scriptsize{± 0.46} &
  \multicolumn{1}{c|}{45.65 \scriptsize{± 0.58}} &
  56.54 \scriptsize{± 0.5} &
  \multicolumn{1}{c|}{{\ul 86.53} \scriptsize{± 0.47}} &
  {\ul 93.24} \scriptsize{± 0.30} &
  \multicolumn{1}{c|}{{\ul 92.05} \scriptsize{± 0.50}} &
  {\ul 96.83} \scriptsize{± 0.27} \\ 
\multicolumn{1}{l|}{ConFeSS \citep{confess}} &
  \multicolumn{1}{c|}{{\ul 27.09}} &
  {\ul 33.57} &
  \multicolumn{1}{c|}{\textbf{48.85}} &
  {\ul 60.10} &
  \multicolumn{1}{c|}{84.65} &
  90.40 &
  \multicolumn{1}{c|}{88.88} &
  95.34 \\
  \midrule
\rowcolor[HTML]{E7F5F8} 
\multicolumn{1}{l|}{\cellcolor[HTML]{E7F5F8}\textbf{\ourmethod{} \textbf{(Ours)}}} &
  \multicolumn{1}{c|}{\cellcolor[HTML]{E7F5F8}\textbf{28.46} \scriptsize{± 0.23}} &
  \textbf{34.21} \scriptsize{± 0.25} &
  \multicolumn{1}{c|}{\cellcolor[HTML]{E7F5F8}44.48 \scriptsize{± 0.31}} &
  56.89 \scriptsize{± 0.29} &
  \multicolumn{1}{c|}{\cellcolor[HTML]{E7F5F8}\textbf{88.55} \scriptsize{± 0.23}} &
  \textbf{93.92} \scriptsize{± 0.14} &
  \multicolumn{1}{c|}{\cellcolor[HTML]{E7F5F8}\textbf{93.65} \scriptsize{± 0.25}} &
  \multicolumn{1}{c}{\cellcolor[HTML]{E7F5F8}\textbf{97.72} \scriptsize{± 0.13}}\\ \bottomrule
\end{tabular}%
}
\vspace{-12pt}
\end{table}

\textbf{$[$A1-b$]$ Cross-Domain Setting.}\label{sssec:cross_domain} 
Following \cite{cdfsl}, we pretrain on miniImageNet and evaluate on CDFSL, the results of which are summarized in Tables~\ref{tab:cdfsl}. \ourmethod{} again sets a new state-of-the-art on ChestX, EuroSAT, and CropDiseases, and remains competitive on ISIC. Notably, the data distributions of ChestX and ISIC are considerably different from that of miniImageNet. We argue that this influences the embedding quality for the downstream inference, and thus, the efficacy of \ourft{} in addressing sample bias. 
Extended versions of Tables~\ref{tab:mini-tiered}-\ref{tab:cdfsl} are found in Appendix~\ref{sec:appendix-addit-experiments}.

\begin{figure}[t]
    \begin{minipage}[t]{0.3\linewidth}
        \begin{overpic}[abs, unit=1cm, width=\columnwidth, clip, percent]{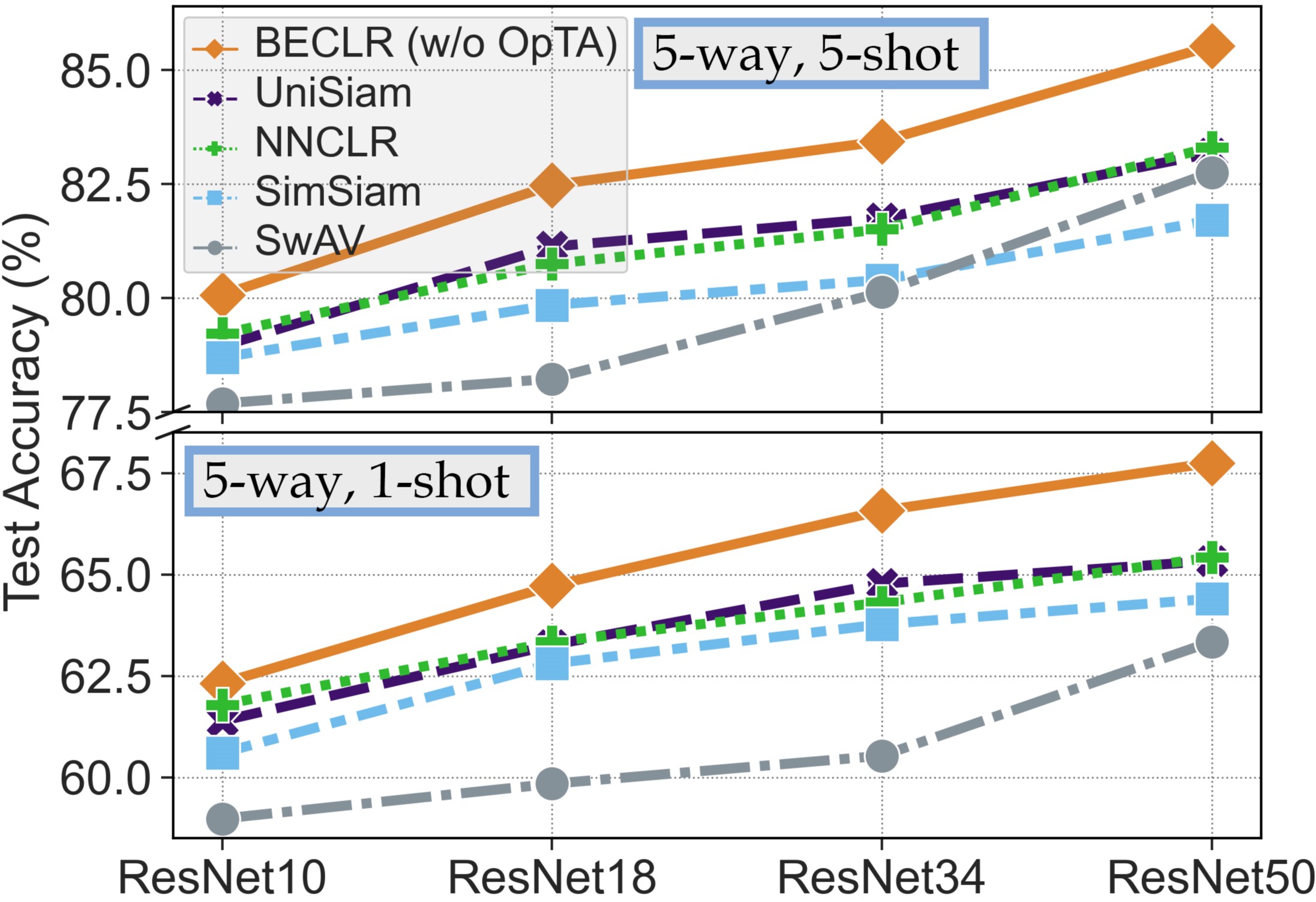}
        \end{overpic}
        \caption{\small \ourmethod{} outperforms all baselines, in terms of U-FSL performance on miniImageNet, even without \ourft{}.}
        \label{fig:bakcbone-study}
    \end{minipage} 
        \hfill%
    \begin{minipage}[t]{0.39\linewidth}
        \includegraphics[width=\linewidth]{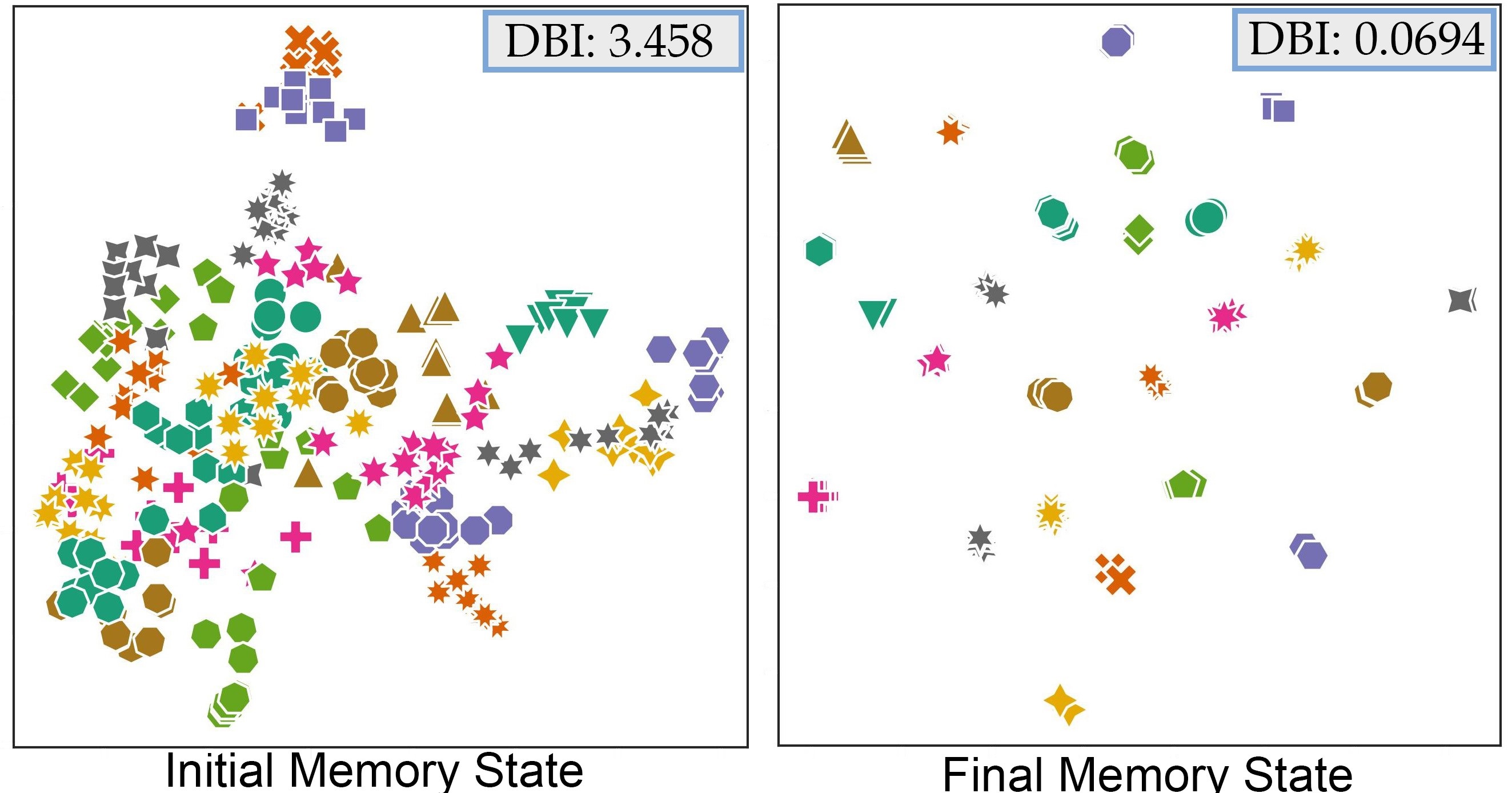}
        \caption{\small The dynamic updates of \ourmodule{} allow the memory to evolve into a highly separable partitioned latent space. Clusters are denoted by different (colors, markers).}
        \label{fig:mem-evol}
    \end{minipage}%
        \hfill%
    \begin{minipage}[t]{0.265\linewidth}
        \includegraphics[width=\linewidth]{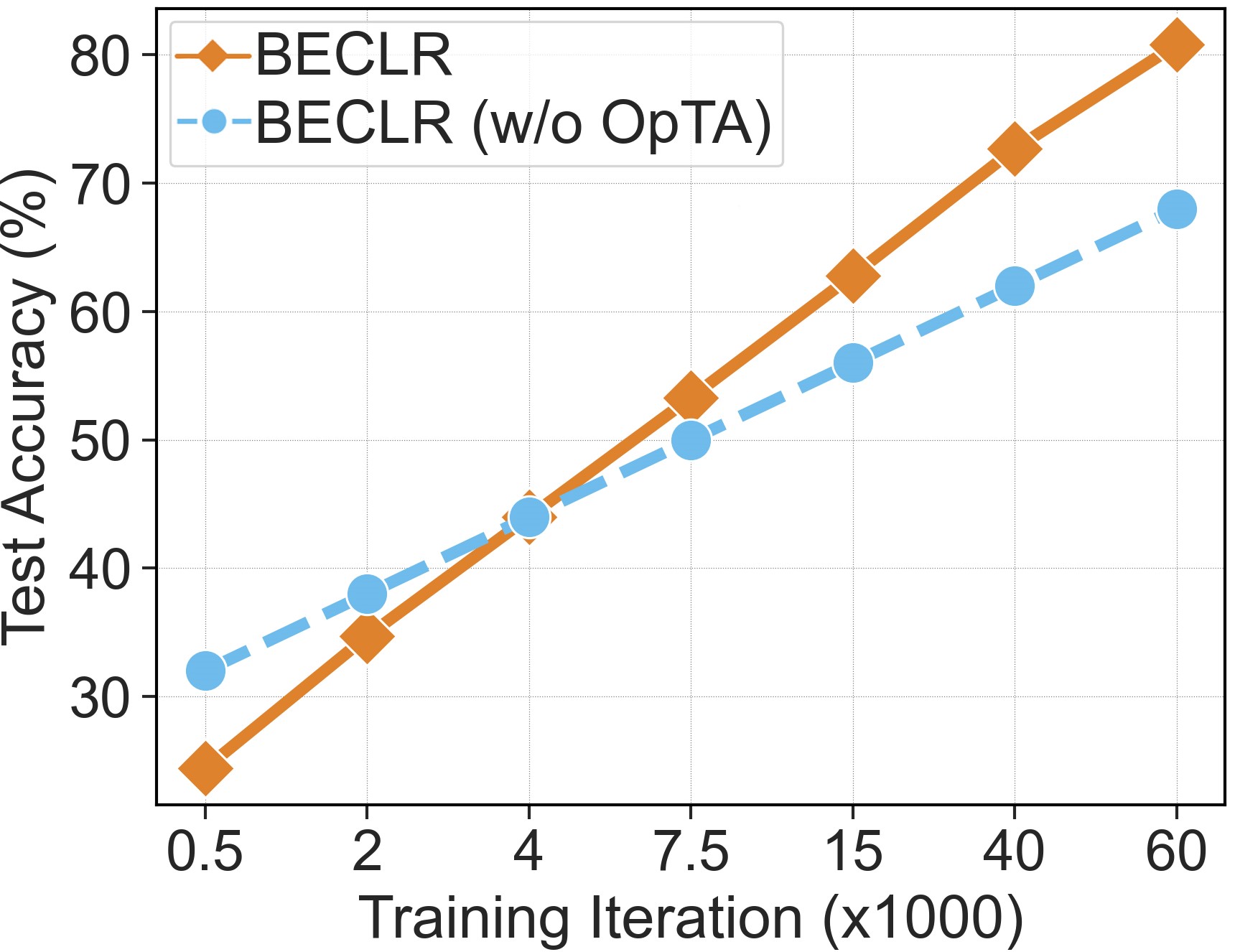}
        \caption{\small 
        \ourft{} produces an increasingly larger performance boost as the pretraining feature quality increases.}
        \label{fig:optaln-gain}
    \end{minipage} 
        \vspace{-0.5cm}
\end{figure}

\begin{wraptable}{r}{0.56\columnwidth}
\centering
\vspace{-12pt}
\aboverulesep = 0pt
\belowrulesep = 0pt
\renewcommand{\arraystretch}{1.2}
\caption{\small \ourmethod{} outperforms enhanced prior art with \ourft{}.}
\label{tab:opta}
\resizebox{0.56\textwidth}{!}{%
\begin{tabular}{@{}l|cc|cc@{}}
\toprule
               & \multicolumn{2}{c}{\cellcolor[HTML]{DCE7D9}\textbf{miniImageNet}} & \multicolumn{2}{c}{\cellcolor[HTML]{FDE5C9}\textbf{tieredImageNet}}  \\ \midrule
\textbf{Method}         & \textbf{1 shot}   & \textbf{5 shot}  & \textbf{1 shot}    & \textbf{5 shot}     \\ \midrule
CPNWCP+\ourft{} \citep{wang2022contrastive}  &  60.45 \scriptsize{± 0.81}    & 75.84 \scriptsize{± 0.56} & 55.05 \scriptsize{± 0.31}     & 72.91 \scriptsize{± 0.26}    \\
HMS+\ourft{} \citep{hms}     &  69.85 \scriptsize{± 0.42}    & 80.77 \scriptsize{± 0.35}     & 71.75  \scriptsize{± 0.43}     & 81.32  \scriptsize{± 0.34} \\
PsCo+\ourft{} \citep{psco}    & 52.89 \scriptsize{± 0.71}    & 67.42 \scriptsize{± 0.54} & 57.46  \scriptsize{± 0.59}     & 70.70  \scriptsize{± 0.45}           \\
UniSiam+\ourft{} \citep{unisiam} & 72.54 \scriptsize{± 0.61}    & 82.46 \scriptsize{± 0.32}    & 73.37 \scriptsize{± 0.64}     & 73.37 \scriptsize{± 0.64} \\ \midrule
\multicolumn{1}{l|}{\cellcolor[HTML]{E7F5F8}\textbf{\ourmethod{} \textbf{(Ours)}}}         & \multicolumn{1}{c}{\cellcolor[HTML]{E7F5F8}\textbf{75.74 \scriptsize{± 0.62}}} & \multicolumn{1}{c}{\cellcolor[HTML]{E7F5F8}\textbf{84.93 \scriptsize{± 0.33}}} & \multicolumn{1}{c}{\cellcolor[HTML]{E7F5F8}\textbf{76.44 \scriptsize{± 0.66}}} & \multicolumn{1}{c}{\cellcolor[HTML]{E7F5F8}\textbf{84.85 \scriptsize{± 0.37}}} \\ \bottomrule
\end{tabular}%
}
\vspace{-12pt}
\end{wraptable}
\textbf{$[$A1-c$]$ Pure Pretraining and \ourft{}.} To substantiate the impact of the design choices in \ourmethod{}, we compare against some of the most influential contrastive SSL approaches: SwAV, SimSiam, NNCLR, and the prior U-FSL state-of-the-art: UniSiam \citep{unisiam}, in terms of pure pretraining performance, by directly evaluating the pretrained model on downstream FSL tasks (i.e., no \ourft{} and no fine-tuning). Fig.~\ref{fig:bakcbone-study} summarizes this comparison for various network depths in the ($5$-way, $\{1,5\}$-shot) settings on miniImageNet. \ourmethod{} again outperforms all U-FSL/SSL frameworks for all backbone configurations, even without \ourft{}.  As an additional study, in Table~\ref{tab:opta} we take the opposite steps by plugging in \ourft{} on a suite of recent prior art in U-FSL. The results demonstrate two important points: (i) \ourft{} is in fact agnostic to the choice of pretraining method, having considerable impact on downstream performance, and (ii) there still exists a margin between enhanced prior art and \ourmethod{}, corroborating that it is not just \ourft{} that has a meaningful effect but also \ourmodule{} and our pretraining methodology.

\textbf{$[$A2$]$ Latent Memory Space Evolution.} As a qualitative demonstration, we visualize $30$ memory embeddings from $25$ partitions $\mathcal{P}_i$ within \ourmodule{} for the initial (left) and final (right) state of the latent memory space ($\mathcal{M}$). The 2-D UMAP plots in Fig.~\ref{fig:mem-evol} provide qualitative evidence of a significant improvement in terms of cluster separation, as training progresses. To quantitatively substantiate this finding, the quality of the memory clusters is also measured by the DBI score \citep{davies1979cluster}, with a lower DBI indicating better inter-cluster separation and intra-cluster tightness. The DBI value is significantly lower between partitions $\mathcal{P}_i$ in the final state 
of $\mathcal{M}$, further corroborating \ourmodule{}'s ability to establish highly separable and meaningful partitions. 
%
\begin{wrapfigure}{r}{0.48\columnwidth}
    \vspace{-12pt}
    \begin{overpic}[abs, unit=1cm, width=0.45\columnwidth, clip, percent]{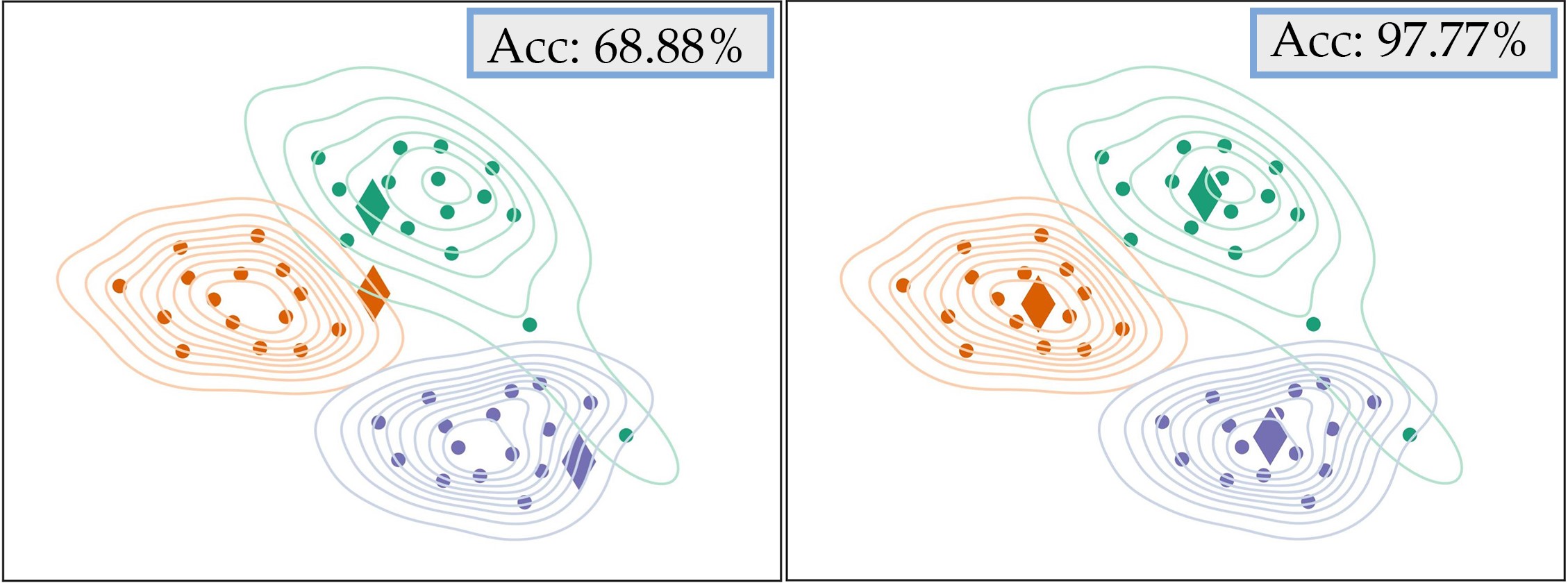}
                \put(2, 1.5){\scriptsize Before \ourft{}}
                \put(52, 1.5){\scriptsize After \ourft{}}
        \end{overpic}
        \caption{\small \ourft{} addresses sample bias, reducing the distribution shift between support and query sets.}
        \label{fig:optaln}
    \vspace{-12pt}
\end{wrapfigure}
\textbf{$[$A3-a$]$ Impact of \ourft.}\label{ssec:impact_OpT-ALN} We visualize the UMAP projections for a randomly sampled ($3$-way, $1$-shot) miniImageNet episode. Fig.~\ref{fig:optaln} illustrates the original $\bm{P}^\mathcal{S}$ (left) and transported $\hat{\bm{P}}^\mathcal{S}$ (right) support prototypes (\scalebox{0.9}{$\blacklozenge$}), along with the query set embeddings $\bm{Z}^\mathcal{Q}$ (\scalebox{0.9}{$\bullet$}) and their kernel density distributions (in contours). As can be seen, the original prototypes are highly biased and deviate from the latent query distributions. \ourft{} pushes the transported prototypes much closer to the query distributions (contour centers), effectively diminishing sample bias, resulting in significantly higher accuracy in the corresponding episode.

\textbf{$[$A3-b$]$ Relation between Pretraining and Inference.}\label{ssec:relation-study} \ourft{} operates under the assumption that the query embeddings $\bm{Z}^\mathcal{Q}$ are representative of the actual class distributions. As such, we argue that its efficiency depends on the quality of the extracted features through the pretrained encoder.  Fig.~\ref{fig:optaln-gain} assesses this hypothesis by comparing pure pretraining (i.e., \ourmethod{} without \ourft{}{}) and downstream performance on miniImageNet for the ($5$-way, $1$-shot) setting as training progresses. As can be seen, when the initial pretraining performance is poor, \ourft{} even leads to performance degradation. On the contrary, it offers an increasingly larger boost as pretraining accuracy improves. The key message here is that these two steps (pretraining and inference) are highly intertwined, further enhancing the overall performance. This notion sits at the core of the \ourmethod{} design strategy.

\vspace{-0.2cm}
\subsection{Ablation Studies}
\label{ssec:ablation-studyies}
\vspace{-0.2cm}
%

%
\begin{wraptable}{r}{0.47\columnwidth}
\vspace{-12pt}
\centering
\caption{Ablating main components of \ourmethod{}.}
\vspace{-0.1cm}
\label{tab:main-abl}
\resizebox{0.47\columnwidth}{!}{%
\begin{tabular}{@{}cccccc@{}}
\toprule
\textbf{Masking} & \textbf{EMA Teacher} & \textbf{\ourmodule{}} & \textbf{\ourft{}} & \textbf{5-way 1-shot}          & \textbf{5-way 5-shot}          \\ \midrule \midrule
-       & -   & -         & -       & 63.57 \scriptsize{± 0.43}          & 81.42 \scriptsize{± 0.28}          \\ \midrule
\cmark       & -   & -         & -       & 54.53 \scriptsize{± 0.42}   & 68.35 \scriptsize{± 0.27}  \\
-       & \cmark   & -         & -       & 65.02 \scriptsize{± 0.41} & 82.33 \scriptsize{± 0.25} \\ \midrule
\cmark       & \cmark   & -         & -       & 65.33 \scriptsize{± 0.44} & 82.69 \scriptsize{± 0.26} \\ \midrule
\cmark       & \cmark   & \cmark         & -       & 67.75 \scriptsize{± 0.43}  & 85.53 \scriptsize{± 0.27} \\ \midrule
\multicolumn{1}{c}{\cellcolor[HTML]{E7F5F8}\cmark}       & \multicolumn{1}{c}{\cellcolor[HTML]{E7F5F8}\cmark}   & \multicolumn{1}{c}{\cellcolor[HTML]{E7F5F8}\cmark}         & \multicolumn{1}{c}{\cellcolor[HTML]{E7F5F8}\cmark}       & \multicolumn{1}{c}{\cellcolor[HTML]{E7F5F8}\textbf{80.57} \scriptsize{± 0.57}} & \multicolumn{1}{c}{\cellcolor[HTML]{E7F5F8}\textbf{87.82} \scriptsize{± 0.29}}   \\ \bottomrule
\end{tabular}%
}
\vspace{-12pt}
\end{wraptable}
\textbf{Main Components of \ourmethod{}.} Let us investigate the impact of sequentially adding in the four main components of \ourmethod{}'s end-to-end architecture: (i) masking, (ii) EMA teacher encoder, (iii) \ourmodule{} and \ourft{}. As can be seen from Table~\ref{tab:main-abl}, when applied individually, masking degrades the performance, but when combined with EMA, it gives a slight boost ($1.5\%$) for both $\{1,5\}$-shot settings. \ourmodule{} and \ourft{} are the most crucial components contributing to the overall performance of \ourmethod{}. \ourmodule{} offers an extra $2.4\%$ and $2.8\%$ accuracy boost in the $1$-shot and $5$-shot settings, respectively, and \ourft{} provides another $12.8\%$ and $2.3\%$ performance gain, in the aforementioned settings. As discussed earlier, also illustrated in Fig.~\ref{fig:optaln-gain}, the gain of \ourft{} is owing and proportional to the performance of \ourmodule{}. This boost is paramount in the $1$-shot scenario where the sample bias is severe.

\begin{table}[t]
\vspace{-5pt}
\centering
\aboverulesep = 0pt
\belowrulesep = 0pt
\renewcommand{\arraystretch}{1.2}
\caption{\small Hyperparameter ablation study for miniImageNet ($5$-way, $5$-shot) tasks. Accuracies in (\% $\pm$ std).}
\vspace{-0.1cm}
\label{tab:ablation}
\resizebox{\textwidth}{!}{%
\begin{tabular}{@{}cc|cc|cc|cc|cc|cc|cc@{}}
\toprule
\multicolumn{2}{c|}{\textbf{Masking Ratio}} &
\multicolumn{2}{c|}{\textbf{Output Dim. ($d$)}} &
\multicolumn{2}{c|}{\textbf{Neg. Loss Weight ($\lambda$)}} &
\multicolumn{2}{c|}{\textbf{\# of NNs ($k$)}} &
\multicolumn{2}{c|}{\textbf{\# of Clusters ($P$)}} &
\multicolumn{2}{c|}{\textbf{Memory Size ($M$)}} &
\multicolumn{2}{c}{\textbf{Memory Module Configuration}} \\ 
\cmidrule(lr){1-2}
\cmidrule(lr){3-4}
\cmidrule(lr){5-6}
\cmidrule(lr){7-8}
\cmidrule(lr){9-10}
\cmidrule(lr){11-12}
\cmidrule(lr){13-14}
\textbf{Value} & \textbf{Accuracy}     & \textbf{Value} & \textbf{Accuracy}     & \textbf{Value} & \textbf{Accuracy}     & \textbf{Value} & \textbf{Accuracy}     & \textbf{Value} & \textbf{Accuracy}     & \textbf{Value} & \textbf{Accuracy}     & \textbf{Value}     & \textbf{Accuracy}     \\ \midrule

\multicolumn{1}{c}{10\%}  & 86.59  \scriptsize{± 0.25} & 256   & 85.16  \scriptsize{± 0.26} & 0.0   & 85.45  \scriptsize{± 0.27} & 1     & 86.58  \scriptsize{± 0.27} & 100   & 85.27  \scriptsize{± 0.24} & 2048  & 85.38  \scriptsize{± 0.25} & \multicolumn{1}{c}{\texttt{DyCE-FIFO}}      

& \multicolumn{1}{c}{84.05  \scriptsize{± 0.39}} \\ \multicolumn{1}{c}{\cellcolor[HTML]{E7F5F8}30\%} & \multicolumn{1}{c|}{\cellcolor[HTML]{E7F5F8}\textbf{87.82} \scriptsize{± 0.29}} & \multicolumn{1}{c}{\cellcolor[HTML]{E7F5F8}512}   & \multicolumn{1}{c|}{\cellcolor[HTML]{E7F5F8}\textbf{87.82} \scriptsize{± 0.29}} & \multicolumn{1}{c}{\cellcolor[HTML]{E7F5F8}0.1}   & \multicolumn{1}{c|}{\cellcolor[HTML]{E7F5F8}\textbf{87.82} \scriptsize{± 0.29}}  & \multicolumn{1}{c}{\cellcolor[HTML]{E7F5F8}3}     & \multicolumn{1}{c|}{\cellcolor[HTML]{E7F5F8}\textbf{87.82} \scriptsize{± 0.29}} & \multicolumn{1}{c}{\cellcolor[HTML]{E7F5F8}200}   & \multicolumn{1}{c|}{\cellcolor[HTML]{E7F5F8}\textbf{87.82} \scriptsize{± 0.29}} & 4096  & 86.28  \scriptsize{± 0.29}  & \multicolumn{1}{c}{\texttt{DyCE-kMeans}} 

& \multicolumn{1}{c}{85.37  \scriptsize{± 0.33}} \\
\multicolumn{1}{c}{50\%}  & 83.36  \scriptsize{± 0.28} & 1024  & 85.93  \scriptsize{± 0.31} & 0.3   & 86.33  \scriptsize{± 0.29}  & 5     & 86.79  \scriptsize{± 0.26} & 300   & 85.81  \scriptsize{± 0.25} & \multicolumn{1}{c}{\cellcolor[HTML]{E7F5F8}8192}  & \multicolumn{1}{c|}{\cellcolor[HTML]{E7F5F8}\textbf{87.82} \scriptsize{± 0.29}}  & \multicolumn{1}{c}{\cellcolor[HTML]{E7F5F8}\ourmodule{}} 

& \multicolumn{1}{c}{\cellcolor[HTML]{E7F5F8} \textbf{87.82} \scriptsize{± 0.29}}\\
\multicolumn{1}{c}{70\%}  & 77.70  \scriptsize{± 0.20} & 2054  & 85.42  \scriptsize{± 0.34} & 0.5   & 85.63  \scriptsize{± 0.26}  & 10    & 86.17 \scriptsize{± 0.28}  & 500   & 85.45  \scriptsize{± 0.20} & 12288 & 85.84  \scriptsize{± 0.22}  &  &              \\ \bottomrule
\end{tabular}%
}
\vspace{-15pt}
\end{table}

\textbf{Other Hyperparameters.} In Table~\ref{tab:ablation}, we summarize the result of ablations on: (i) the masking ratio of student images, (ii) the embedding latent dimension $d$, (iii) the loss weighting hyperparameter $\lambda$ (in Eq.~\ref{eqn:finalloss}), and regarding \ourmodule{}: (iv) the number of nearest neighbors selected $k$, (v) the number of memory partitions/clusters $P$, (vi) the size of the memory $M$, and (vii) different memory module configurations. As can be seen, a random masking ratio of $30\%$ yields the best performance. We find that $d = 512$ gives the best results, which is consistent with the literature \citep{byol,simsiam}. The negative loss term turns out to be unnecessary to prevent representation collapse (as can be seen when $\lambda = 0$) and $\lambda=0.1$ yields the best performance. Regarding the number of neighbors selected ($k$), there seems to be a sweet spot in this setting around $k = 3$, where increasing further would lead to inclusion of potentially false positives, and thus performance degradation. $P$ and $M$ are tunable hyperparameters (per dataset) that return the best performance at $P=200$ and $M=8192$ in this setting. Increasing $P$, $M$ beyond a certain threshold appears to have a negative impact on cluster formation. We argue that extremely large memory would result in accumulating old embeddings, which might no longer be a good representative of their corresponding classes. Finally, we compare \ourmodule{} against two \emph{degenerate} versions of itself: \texttt{DyCE-FIFO}, where the memory has no partitions and is updated with a first-in-first-out strategy; here for incoming embeddings we pick the closest $k$ neighbors. \texttt{DyCE-kMeans}, where we preserve the memory structure and only replace optimal transport with $\operatorname{kMeans}$ (with $P$ clusters), in line $8$ of Algorithm~\ref{alg:DyCE-ver1}. The performance drop in both cases confirms the importance of the proposed mechanics within \ourmodule{}. 

\vspace{-0.2cm}
\section{Concluding Remarks and Broader Impact}
\label{sec:conclusion}
\vspace{-0.2cm}
In this paper, we have articulated two key shortcomings of the prior art in U-FSL, to address each of which we have proposed a novel solution embedded within the proposed end-to-end approach, \ourmethod{}. The first angle of novelty in \ourmethod{} is its dynamic clustered memory module (coined as \ourmodule{}) to enhance positive sampling in contrastive learning. The second angle of novelty is an efficient distribution alignment strategy (called \ourft{}) to address the inherent sample bias in (U-)FSL. Even though tailored towards U-FSL, we believe \ourmodule{} has potential \emph{broader impact} on generic self-supervised learning state-of-the-art, as we already demonstrate (in Section~\ref{sec:experimental-eval}) that even with \ourft{} unplugged, \ourmodule{} alone empowers \ourmethod{} to still outperform the likes of SwaV, SimSiam and NNCLR. 
\ourft{}, on the other hand, is an efficient add-on module, which we argue has to become an \emph{integral part} of all (U-)FSL approaches, especially in the more challenging low-shot scenarios.

\newpage
\section*{Reproducibility Statement.} To help readers reproduce our experiments, we provide extensive descriptions of implementation details and algorithms. Architectural and training details are provided in Appendices~\ref{ssec:appendix-arch} and~\ref{ssec:appendix-train}, respectively, along with information on the applied data augmentations (Appendix~\ref{ssec:appendix-augm}) and tested benchmark datasets (Appendix~\ref{ssec:appendix-datasets}). The algorithms for \ourmethod{} pretraining and \ourmodule{} are also provided in both algorithmic (in Algorithms~\ref{alg:BECLR-ver1}, \ref{alg:DyCE-ver1}) and Pytorch-like pseudocode formats (in Algorithms~\ref{alg:BECLR}, \ref{alg:DyCE}). We have taken every measure to ensure fairness in our comparisons by following the most commonly adopted pretraining and evaluation settings in the U-FSL literature in terms of: pretraining/inference benchmark datasets used for both in-domain and cross-domain experiments, pretraining data augmentations, ($N$-way, $K$-shot) inference settings and number of query set images per tested episode. We also draw baseline results from their corresponding original papers and compare their performance with \ourmethod{} for identical backbone depths. For our reproductions (denoted as $^\dagger$) of SwAV and NNCLR we follow the codebase of the original work and adopt it in the U-FSL setup, by following the same augmentations, backbones, and evaluation settings as \ourmethod{}. Our codebase is also provided as part of our supplementary material, in an anonymized fashion, and will be made publicly available upon acceptance. All training and evaluation experiments are conducted on $2$ A40 NVIDIA GPUs.

\section*{Ethics Statement.} We have read the ICLR Code of Ethics (\MYhref[darkcyan]{https://iclr.cc/public/CodeOfEthics}{https://iclr.cc/public/CodeOfEthics}) and ensured that this work adheres to it. All benchmark datasets and pretrained model checkpoints are publicly available and not directly subject to ethical concerns.

\section*{Acknowledgements.} The authors would like to thank Marcel Reinders, Ojas Shirekar and Holger Caesar for insightful conversations and brainstorming. This work was in part supported by \MYhref[darkcyan]{https://www.shell.com/energy-and-innovation/digitalisation/shell-ai.html}{Shell.ai} Innovation program.

\bibliography{ref}
\bibliographystyle{iclr2024_conference}

\appendix
\clearpage

\section{Implementation Details}
\label{sec:appendix-impl-det}
\vspace{-0.2cm}
This section describes the implementation and training details of \ourmethod{}.

\subsection{Architecture Details}
\label{ssec:appendix-arch}
\vspace{-0.2cm}

\ourmethod{} is implemented on PyTorch \citep{paszke2019pytorch}. We use the ResNet family \citep{resnets} for our backbone networks ($f_\theta$, $f_\psi$). The projection ($g_\theta$, $g_\psi$) and prediction ($h_\theta$) heads are $3$- and $2$-layer MLPs, respectively, as in \citet{simsiam}. Batch normalization (BN) and a ReLU activation function are applied to each MLP layer, except for the output layers. No ReLU activation is applied on the output layer of the projection heads ($g_\theta$, $g_\psi$), while neither BN nor a ReLU activation is applied on the output layer of the prediction head ($h_\theta$). We use a resolution of $224 \times 224$ for input images and a latent embedding dimension of $d = 512$ in all models and experiments, unless otherwise stated. The \ourmodule{} memory module consists of a memory unit $\mathcal{M}$, initialized as a random table (of size $M \times d$). We also maintain up to $P$ partitions in $\mathcal{M}=[\mathcal{P}_1,\ldots,\mathcal{P}_P]$, each of which is represented by a prototype stored in $\bm{\Gamma}=[\bm{\gamma}_1,\ldots,\bm{\gamma}_P]$. Prototypes $\bm{\gamma}_i$ are the average of the latent embeddings stored in partition $\mathcal{P}_i$. When training on miniImageNet, CIFAR-FS and FC$100$, we use a memory of size $M = 8192$ that contains $P = 200$ partitions and cluster prototypes ($M = 40960$, $P = 1000$ for tieredImageNet). Note that both $M$ and $P$ are important hyperparameters, whose values were selected by evaluating on the validation set of each dataset for model selection. These hyperparameters would also need to be carefully tuned on an unknown unlabeled training dataset.

\subsection{Training Details}
\label{ssec:appendix-train}
\vspace{-0.2cm}

\ourmethod{} is pretrained on the training splits of miniImageNet, CIFAR-FS, FC$100$ and tieredImageNet. We use a batch size of $B = 256$ images for all datasets, except for tieredImageNet ($B=512$). Following \citet{simsiam}, images are resized to $224 \times 224$ for all configurations. We use the SGD optimizer with a weight decay of $10^{-4}$, a momentum of $0.995$, and a cosine decay schedule of the learning rate. Note that we do not require large-batch optimizers, such as LARS \citep{you2017large}, or early stopping. Similarly to \citet{unisiam}, the initial learning rate is set to $0.3$ for the smaller miniImageNet, CIFAR-FS, FC$100$ datasets and $0.1$ for tieredImageNet, and we train for $400$ and $200$ epochs, respectively. The temperature scalar in the loss function is set to $\tau=2$. Upon finishing unsupervised pretraining, we only keep the last epoch checkpoint of the student encoder ($f_\theta$) for the subsequent inference stage. For the inference and downstream few-shot classification stage, we create ($N$-way, $K$-shot) tasks from the validation and testing splits of each dataset for model selection and evaluation, respectively. In the inference stage, we sequentially perform up to $\delta \leq 5$ consecutive passes of \ourft{}, with the transported prototypes of each pass acting as the input of the next pass. The optimal value for $\delta$ for each dataset and ($N$-way, $K$-shot) setting is selected by evaluating on the validation dataset. 

Note that at the beginning of pretraining both the encoder representations and the memory embedding space within \ourmodule{} are highly volatile. Thus, we allow for an adaptation period $\operatorname{epoch} < \operatorname{epoch}_\textup{\texttt{thr}}$ (empirically $20$-$50$ epochs), during which the batch enhancement path of \ourmodule{} is not activated and the encoder is trained via standard contrastive learning (without enhancing the batch with additional positives). On the contrary, the memory updating path of \ourmodule{} is activated for every training batch from the beginning of training, allowing the memory to reach a highly separable converged state (see Fig.~\ref{fig:mem-evol}), before plugging in the batch enhancement path in the \ourmethod{} pipeline. When the memory space ($\mathcal{M}$) is full for the first time, a kmeans \citep{kmeans} clustering step is performed for initializing the cluster prototypes ($\bm{\gamma}_i$) and memory partitions ($\mathcal{P}_i$). This kmeans clustering step is performed only once during training to initialize the memory prototypes, which are then dynamically updated for each training batch by the memory updating path of \ourmodule{}.

\subsection{Image Augmentations}
\label{ssec:appendix-augm}
\vspace{-0.2cm}

The data augmentations that were applied in the pretraining stage of \ourmethod{} are showcased in Table~\ref{tab:data-augmentations}. These augmentations were applied on the input images for all training datasets. The default data augmentation profile follows a common data augmentation strategy in SSL, including RandomResizedCrop (with scale in [0.2, 1.0]), random ColorJitter \citep{wu2018unsupervised} of \{brightness, contrast, saturation, hue\} with a probability of $0.1$, RandomGrayScale with a probability of $0.2$, random GaussianBlur with a probability of $0.5$ and a Gaussian kernel in $\left[0.1, 2.0\right]$, and finally, RandomHorizontalFlip with a probability of $0.5$. Following \citet{unisiam}, this profile can be expanded to the strong data augmentation profile, which also includes RadomVerticalFlip with a probability of $0.5$ and RandAugment \citep{cubuk2020randaugment} with $n=2$ layers, a magnitude of $m=10$, and a noise of the standard deviation of magnitude of $mstd=0.5$. Unless otherwise stated, the strong data augmentation profile is applied on all training images before being passed to the backbone encoders.

\begin{table}[t]
\centering
\renewcommand{\arraystretch}{1.1}
\caption{\small Pytorch-like descriptions of the data augmentation profiles applied on the pretraining phase of \ourmethod{}.}
\vspace{-0.1cm}
\label{tab:data-augmentations}
\resizebox{\textwidth}{!}{%
\begin{tabular}{@{}cl@{}}
\toprule
\textbf{Data Augmentation Profile} & \multicolumn{1}{c}{\textbf{Description}} \\ \midrule \midrule
                           & RandomResizedCrop(size=224, scale=(0.2, 1))                                        \\
                           & RandomApply([ColorJitter(brightness=0.4, contrast=0.4, saturation=0.4, hue=0.1)], p=0.1)                                        \\
\textbf{Default}                      & RandomGrayScale(p=0.2)                                         \\
                           & RandomApply([GaussianBlur([0.1, 2.0])], p=0.5)                                        \\
                           & RandomHorizontalFlip(p=0.5)                                        \\ \midrule
                           & RandomResizedCrop(size=224, scale=(0.2, 1))                                        \\
                           & RandomApply([ColorJitter(brightness=0.4, contrast=0.4, saturation=0.2, hue=0.1)], p=0.1)                                        \\
                           & RandomGrayScale(p=0.2)                                         \\
\textbf{Strong}                     &  RandomApply([GaussianBlur([0.1, 2.0])], p=0.5)                                       \\
                           & RandAugment(n=2, m=10, mstd=0.5)                                       \\
                           & RandomHorizontalFlip(p=0.5)                                        \\
\textbf{}                  & RandomVerticalFlip(p=0.5)                                        \\ \bottomrule
\end{tabular}%
}
\end{table}

\section{Experimental Setup}
\label{sec:appendix-exp-setup}
\vspace{-0.2cm}
In this section, we provide more detailed information regarding all the few-shot benchmark datasets that were used as part of our experimental evaluation (in Section~\ref{sec:experimental-eval}), along with \ourmethod{}'s training and evaluation procedures for both in-domain and cross-domain U-FSL settings, for ensuring fair comparisons with U-FSL baselines and our reproductions (for SwAV and NNCLR). 

\begin{table}[h]
\centering
\caption{Overview of cross-domain few-shot benchmarks, on which \ourmethod{} is evaluated. The datasets are sorted with decreasing (distribution) domain similarity to ImageNet and miniImageNet.}
\vspace{-0.1cm}
\label{tab:cd-fsl-datasets}
\resizebox{0.75\textwidth}{!}{%
\begin{tabular}{@{}clcc@{}}
\toprule
\textbf{ImageNet similarity} & \textbf{Dataset} & \textbf{\# of classes} & \textbf{\# of samples} \\ \midrule \midrule
High & CUB \citep{cub}          & 200 & 11,788 \\ \midrule
Low  & CropDiseases \citep{cropdisease} & 38  & 43,456 \\
Low  & EuroSAT \citep{eurosat}      & 10  & 27,000 \\
Low  & ISIC \citep{isic}         & 7   & 10,015 \\
Low  & ChestX \citep{chestx}       & 7   & 25,848 \\ \bottomrule
\end{tabular}%
}
\end{table}

\subsection{Dataset Details}\label{ssec:appendix-data}
\label{ssec:appendix-datasets}
\vspace{-0.2cm}

\textbf{miniImageNet.} It is a subset of ImageNet \citep{imageNet}, containing $100$ classes with $600$ images per class. We randomly select $64$, $16$, and $20$ classes for training, validation, and testing, following the predominantly adopted settings of \citet{ravi2016optimization}. 

\textbf{tieredImageNet.} It is a larger subset of ImageNet, containing $608$ classes and a total of $779,165$ images, grouped into $34$ high-level categories, $20$ ($351$ classes) of which are used for training, $6$ ($97$ classes) for validation and $8$ ($160$ classes) for testing. 

\textbf{CIFAR-FS.} It is a subset of CIFAR-$100$ \citep{cifar}, which is focused on FSL tasks by following the same sampling criteria that were used for creating miniImageNet. It contains $100$ classes with $600$ images per class, grouped into $64$, $16$, $20$ classes for training, validation, and testing, respectively. The additional challenge here is the limited original image resolution of $32 \times 32$.

\textbf{FC100.} It is also a subset of CIFAR-$100$ \citep{cifar} and contains the same $60000$ ($32 \times 32$) images as CIFAR-FS. Here, the original $100$ classes are grouped into $20$ superclasses, in such a way as to minimize the information overlap between training, validation and testing classes \citep{mcallester2020formal}. This makes this data set more demanding for (U-)FSL, since training and testing classes are highly dissimilar. The training split contains $12$ superclasses (of $60$ classes), while both the validation and testing splits are composed of $4$
superclasses (of $20$ classes).

\textbf{CDFSL.} It consists of four distinct datasets with decreasing domain similarity to ImageNet, and by extension miniImageNet, ranging from crop disease images in CropDiseases \citep{cropdisease} and aerial satellite images in EuroSAT \citep{eurosat} to dermatological skin lesion images in ISIC2018 \citep{isic} and grayscale chest X-ray images in ChestX \citep{chestx}. 

\textbf{CUB.} It consists of $200$ classes and a total of $11, 788$ images, split into $100$ classes for training and $50$ for both validation and testing, following the split settings of \citet{chen2019closer}. Additional information on the cross-domain few-shot benchmarks (CUB and CDFSL) is provided in Table~\ref{tab:cd-fsl-datasets}.

\subsection{Pretraining and Evaluation Procedures} 
\vspace{-0.2cm}
For all in-domain experiments \ourmethod{} is pretrained on the training split of the selected dataset (miniImageNet, tieredImageNet, CIFAR-FS, or FC$100$), followed by the subsequent inference stage on the validation and testing splits of the same dataset for model selection and final evaluation, respectively. In contrast, in the cross-domain setting \ourmethod{} is pretrained on the training split of miniImageNet and then evaluated on the validation and test splits of either CDFSL (ChestX, ISIC, EuroSAT, CropDiseases) or CUB. We report test accuracies with $95\%$ confidence intervals over $2000$ test episodes, each with $15$ query shots per class, for all tested datasets, as is most commonly adopted in the literature \citep{pdanet,deepeigenmaps,unisiam}. The performance on miniImageNet, tieredImageNet, CIFAR-FS, FC$100$ and miniImageNet $\rightarrow$ CUB is evaluated on ($5$-way, $\{1,5\}$-shot) classification tasks, while on miniImageNet $\rightarrow$ CDFSL we evaluate on ($5$-way, $\{5,20\}$-shot) tasks, as is customary across the literature \citep{cdfsl}.

We have taken every measure to ensure fairness in our comparison with U-FSL baselines and our reproductions. To do so, all compared baselines have the same pretraining and testing dataset (in both in-domain and cross-domain scenarios), follow similar data augmentation profiles as part of their pretraining and are evaluated in identical ($N$-way, $K$-shot) FSL settings, on the same number of query set images, for all tested inference episodes. We also draw baseline results from their corresponding original papers and compare their performance with \ourmethod{} for identical backbone depths (roughly similar parameter count for all methods). For our reproductions (SwAV and NNCLR), we follow the codebase of the original work and adopt it in the U-FSL setup, by following the same augmentations, backbones, and evaluation settings as \ourmethod{}.

\section{Extended Results and Visualizations}\label{sec:appendix-addit-experiments}
\vspace{-0.2cm}
This section provides extended experimental findings, both quantitative and qualitative, complementing the experimental evaluation in Section~\ref{sec:experimental-eval} and providing further intuition on \ourmethod{}.

\begin{figure}[h]
     \centering
     \begin{subfigure}[b]{0.31\textwidth}
         \centering
         \includegraphics[width=\textwidth]{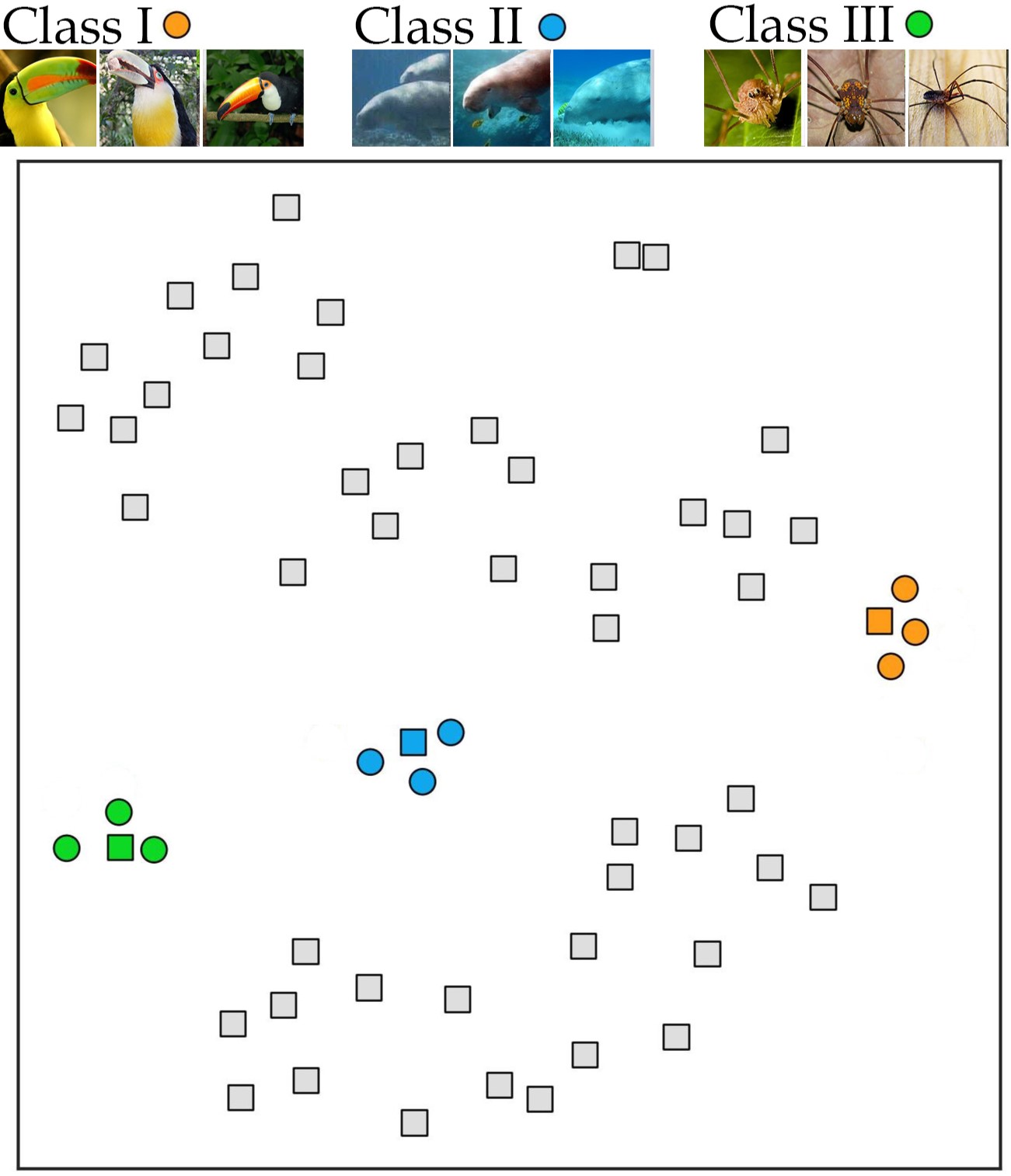}
         \caption{}
         \label{fig:mem-anal-a}
     \end{subfigure}
     \hfill
     \begin{subfigure}[b]{0.31\textwidth}
         \centering
         \includegraphics[width=\textwidth]{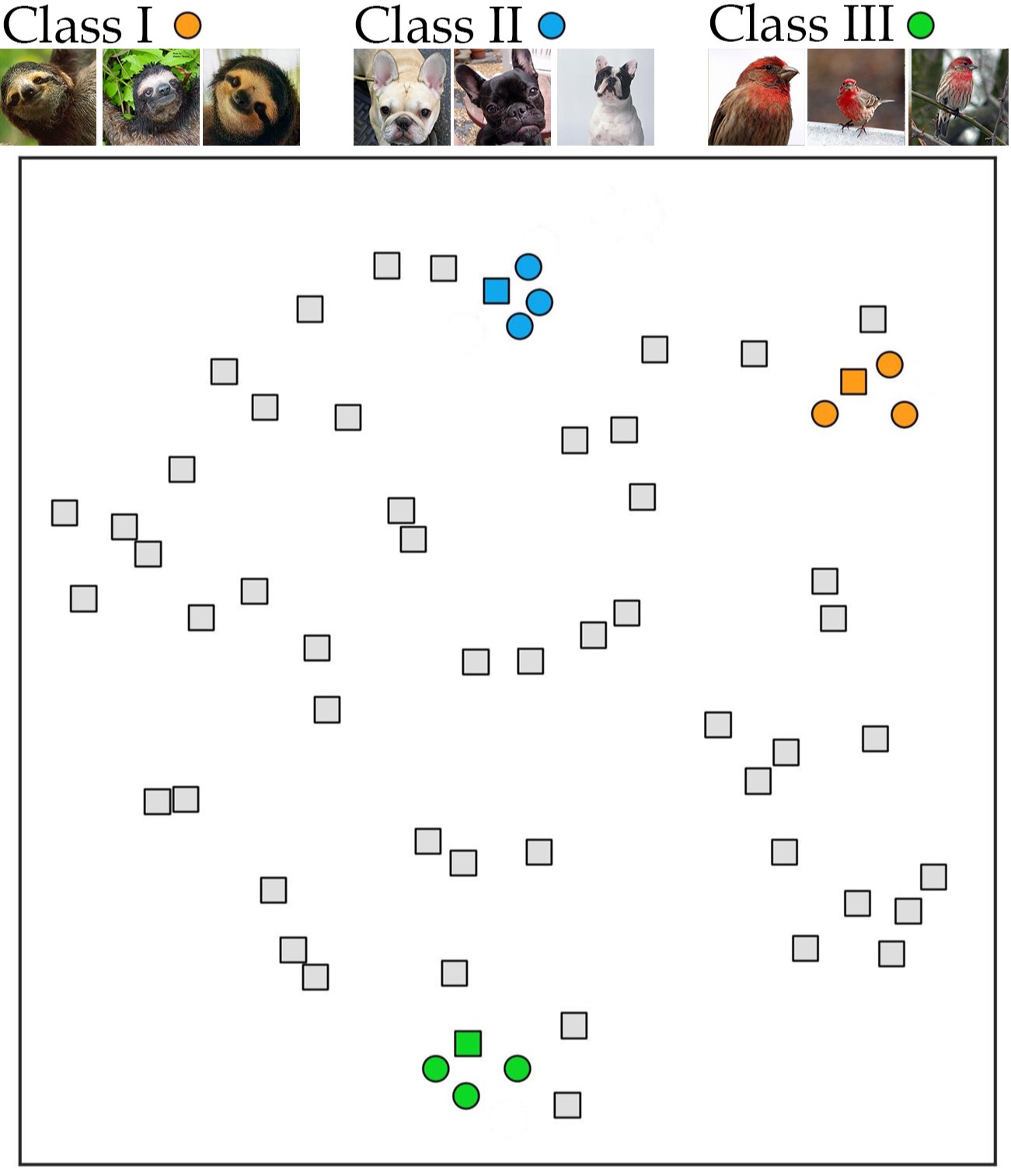}
         \caption{}
         \label{fig:mem-anal-b}
     \end{subfigure}
     \hfill
     \begin{subfigure}[b]{0.31\textwidth}
         \centering
         \includegraphics[width=\textwidth]{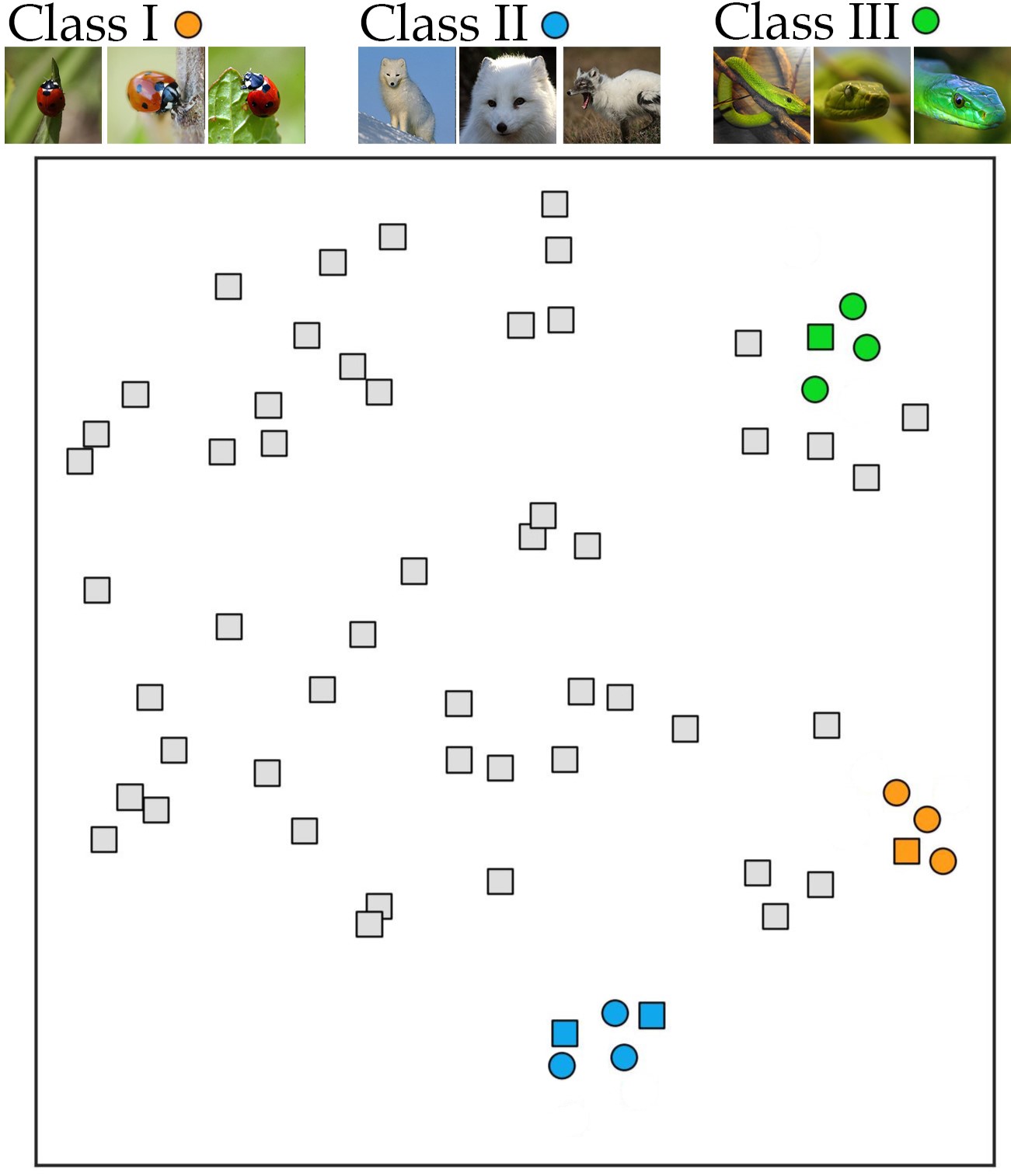}
         \caption{}
         \label{fig:mem-anal-c}
     \end{subfigure}
    \caption{\small (a, b): All $3$ support images ($\bullet$), for all $3$ classes, get assigned to a single class-representative memory prototype (\scalebox{0.7}{$\blacksquare$}). (c): A cluster can be split into multiple subclusters, without degrading the model performance. Colors denote the classes of the random ($3$-way, $3$-shot) episode and the assigned memory prototypes.}
    \label{fig:mem-anal}
    \vspace{-0.2cm}
\end{figure}

\subsection{Analysis on Pseudoclass-Cognizance}
\label{ssec:mem-additional-exp}

\vspace{-0.2cm}
As discussed in Section~\ref{sec:experimental-eval}, the dynamic equipartitioned updates of \ourmodule{} facilitate the creation of a highly separable latent memory space, which is then used for sampling additional meaningful positive pairs. We argue that these separable clusters and their prototypes capture a semblance of class-cognizance. As a qualitative demonstration, Fig.~\ref{fig:mem-anal} depicts the 2-D UMAP \citep{mcinnes2018umap} projections of the support embeddings ($\bullet$) for $3$ (Figs.~\ref{fig:mem-anal-a},~\ref{fig:mem-anal-b},~\ref{fig:mem-anal-c}) randomly sampled miniImageNet ($3$-way, $3$-shot) tasks, along with $50$ randomly selected memory prototypes (\scalebox{0.7}{$\square$}) from the last-epoch checkpoint of \ourmodule{}. Only memory prototypes assigned to support embeddings (\scalebox{0.7}{$\blacksquare$}) are colored (with the corresponding class color). As can be seen in Figs.~\ref{fig:mem-anal-a},~\ref{fig:mem-anal-b}, all $3$-shot embeddings for each class are assigned to a single memory prototype, which shows that the corresponding prototype, and by extension memory cluster, is capable of capturing the latent representation of a class even without access to base class labels. Note that as the training progresses, a memory cluster can split into two or more (sub-)clusters, corresponding to the same latent training class. For example, Fig.~\ref{fig:mem-anal-c} illustrates such a case where the fox class has been split between two different prototypes. This is to be expected since there is no ``one-to-one'' relationship between memory clusters and latent training classes (we do not assume any prior knowledge on the classes of the unlabeled pretraining dataset and the number of memory clusters is an important hyperparameter to be tuned). Notably, \ourmethod{} still treats images sampled from these two (sub-)clusters as positives, which does not degrade the model's performance assuming that both of these (sub-)clusters are consistent (i.e., contain only fox image embeddings).

\begin{table}[th!]
\centering
\aboverulesep = 0pt
\belowrulesep = 0pt
\renewcommand{\arraystretch}{1.2}
\caption{\small Extended version of Table~\ref{tab:mini-tiered}. Accuracies (in \% $\pm$ std) on miniImageNet and tieredImageNet compared against unsupervised (\texttt{Unsup.}) and supervised (\texttt{Sup.}) baselines. Encoders: \texttt{RN}: ResNet, \texttt{Conv}: convolutional blocks. $^\dagger$: denotes our reproduction. $^*$: extra synthetic training data. Style: \textbf{best} and \underline{second best}.}
\vspace{-0.1cm}
\label{tab:mini-tiered-full}
\resizebox{\textwidth}{!}{%
\begin{tabular}{@{}lcccccc@{}}
\toprule
 &
   &
  \multicolumn{1}{c|}{} &
  \multicolumn{2}{c|}{\cellcolor[HTML]{DCE7D9}\textbf{miniImageNet}} &
  \multicolumn{2}{c}{\cellcolor[HTML]{FDE5C9}\textbf{tieredImageNet}} \\ \midrule
\multicolumn{1}{l|}{\textbf{Method}} &
  \multicolumn{1}{c|}{\ \  \ \textbf{Backbone}\ \  \ } &
  \multicolumn{1}{c|}{\textbf{Setting}} &
  \multicolumn{1}{c|}{\ \  \ \textbf{5-way 1-shot}\ \ \ } &
  \multicolumn{1}{c|}{\ \ \ \textbf{5-way 5-shot}\ \ \ } &
  \multicolumn{1}{c|}{\ \ \ \textbf{5-way 1-shot}\ \ \ } &
  \ \ \ \textbf{5-way 5-shot}\ \ \  \\ \midrule
\multicolumn{1}{l|}{ProtoTransfer \citep{prototransfer}} &
  \multicolumn{1}{c|}{$\texttt{Conv4}$} &
  \multicolumn{1}{c|}{$\texttt{Unsup.}$} &
  \multicolumn{1}{c|}{45.67 \scriptsize{± 0.79}} &
  \multicolumn{1}{c|}{62.99 \scriptsize{± 0.75}} &
  \multicolumn{1}{c|}{-} &
  - \\
\multicolumn{1}{l|}{Meta-GMVAE \citep{meta_gmvae}} &
  \multicolumn{1}{c|}{$\texttt{Conv4}$} &
  \multicolumn{1}{c|}{$\texttt{Unsup.}$} &
  \multicolumn{1}{c|}{42.82 \scriptsize{± 0.45}} &
  \multicolumn{1}{c|}{55.73 \scriptsize{± 0.39}} &
  \multicolumn{1}{c|}{-} &
  - \\
\multicolumn{1}{l|}{C3LR \citep{c3lr}} &
  \multicolumn{1}{c|}{$\texttt{Conv4}$} &
  \multicolumn{1}{c|}{$\texttt{Unsup.}$} &
  \multicolumn{1}{c|}{47.92 \scriptsize{± 1.20}} &
  \multicolumn{1}{c|}{64.81 \scriptsize{± 1.15}} &
  \multicolumn{1}{c|}{42.37 \scriptsize{± 0.77}} &
  61.77 \scriptsize{± 0.25} \\
\multicolumn{1}{l|}{SAMPTransfer \citep{samptransfer}} &
  \multicolumn{1}{c|}{$\texttt{Conv4b}$} &
  \multicolumn{1}{c|}{$\texttt{Unsup.}$} &
  \multicolumn{1}{c|}{61.02 \scriptsize{± 1.05}} &
  \multicolumn{1}{c|}{72.52 \scriptsize{± 0.68}} &
  \multicolumn{1}{c|}{49.10 \scriptsize{± 0.94}} &
  65.19 \scriptsize{± 0.82} \\
\multicolumn{1}{l|}{LF2CS \citep{lf2cs}} &
  \multicolumn{1}{c|}{$\texttt{RN12}$} &
  \multicolumn{1}{c|}{$\texttt{Unsup.}$} &
  \multicolumn{1}{c|}{47.93 \scriptsize{± 0.19}} &
  \multicolumn{1}{c|}{66.44 \scriptsize{± 0.17}} &
  \multicolumn{1}{c|}{53.16 \scriptsize{± 0.66}} &
  66.59 \scriptsize{± 0.57} \\
\multicolumn{1}{l|}{CPNWCP \citep{wang2022contrastive}} &
  \multicolumn{1}{c|}{$\texttt{RN18}$} &
  \multicolumn{1}{c|}{$\texttt{Unsup.}$} &
  \multicolumn{1}{c|}{53.14 \scriptsize{± 0.62}} &
  \multicolumn{1}{c|}{67.36 \scriptsize{± 0.5}} &
  \multicolumn{1}{c|}{45.00 \scriptsize{± 0.19}} &
  62.96 \scriptsize{± 0.19} \\
\multicolumn{1}{l|}{SimCLR \citep{simclr}} &
  \multicolumn{1}{c|}{$\texttt{RN18}$} &
  \multicolumn{1}{c|}{$\texttt{Unsup.}$} &
  \multicolumn{1}{c|}{62.58 \scriptsize{± 0.37}} &
  \multicolumn{1}{c|}{79.66 \scriptsize{± 0.27}} &
  \multicolumn{1}{c|}{63.38 \scriptsize{± 0.42}} &
  79.17 \scriptsize{± 0.34} \\
  \multicolumn{1}{l|}{SwAV$^\dagger$ \citep{SwAV}} &
  \multicolumn{1}{c|}{$\texttt{RN18}$} &
  \multicolumn{1}{c|}{$\texttt{Unsup.}$} &
  \multicolumn{1}{c|}{59.84 \scriptsize{± 0.52}} &
  \multicolumn{1}{c|}{78.23 \scriptsize{± 0.26}} &
  \multicolumn{1}{c|}{65.26 \scriptsize{± 0.53}} &
  81.73 \scriptsize{± 0.24} \\
\multicolumn{1}{l|}{NNCLR$^\dagger$ \citep{nnclr}} &
  \multicolumn{1}{c|}{$\texttt{RN18}$} &
  \multicolumn{1}{c|}{$\texttt{Unsup.}$} &
  \multicolumn{1}{c|}{63.33 \scriptsize{± 0.53}} &
  \multicolumn{1}{c|}{80.75 \scriptsize{± 0.25}} &
  \multicolumn{1}{c|}{65.46 \scriptsize{± 0.55}} &
  81.40 \scriptsize{± 0.27} \\
\multicolumn{1}{l|}{SimSiam \citep{simsiam}} &
  \multicolumn{1}{c|}{$\texttt{RN18}$} &
  \multicolumn{1}{c|}{$\texttt{Unsup.}$} &
  \multicolumn{1}{c|}{62.80 \scriptsize{± 0.37}} &
  \multicolumn{1}{c|}{79.85 \scriptsize{± 0.27}} &
  \multicolumn{1}{c|}{64.05 \scriptsize{± 0.40}} &
  81.40 \scriptsize{± 0.30} \\
\multicolumn{1}{l|}{HMS \citep{hms}} &
  \multicolumn{1}{c|}{$\texttt{RN18}$} &
  \multicolumn{1}{c|}{$\texttt{Unsup.}$} &
  \multicolumn{1}{c|}{58.20 \scriptsize{± 0.23}} &
  \multicolumn{1}{c|}{75.77 \scriptsize{± 0.16}} &
  \multicolumn{1}{c|}{58.42 \scriptsize{± 0.25}} &
  75.85 \scriptsize{± 0.18} \\
\multicolumn{1}{l|}{Laplacian Eigenmaps \citep{deepeigenmaps}} &
  \multicolumn{1}{c|}{$\texttt{RN18}$} &
  \multicolumn{1}{c|}{$\texttt{Unsup.}$} &
  \multicolumn{1}{c|}{59.47 \scriptsize{± 0.87}} &
  \multicolumn{1}{c|}{78.79 \scriptsize{± 0.58}} &
  \multicolumn{1}{c|}{-} &
  - \\
\multicolumn{1}{l|}{\textcolor{black}{PsCo}$^\dagger$ \citep{psco}} &
  \multicolumn{1}{c|}{\textcolor{black}{$\texttt{RN18}$}} &
  \multicolumn{1}{c|}{\textcolor{black}{$\texttt{Unsup.}$}} &
  \multicolumn{1}{c|}{\textcolor{black}{47.24 \scriptsize{± 0.76}}} &
  \multicolumn{1}{c|}{\textcolor{black}{65.48 \scriptsize{± 0.68}}} &
  \multicolumn{1}{c|}{\textcolor{black}{54.33 \scriptsize{± 0.54}}} &
  \textcolor{black}{69.73 \scriptsize{± 0.49}} \\
\multicolumn{1}{l|}{UniSiam + dist \citep{unisiam}} &
  \multicolumn{1}{c|}{$\texttt{RN18}$} &
  \multicolumn{1}{c|}{$\texttt{Unsup.}$} &
  \multicolumn{1}{c|}{64.10 \scriptsize{± 0.36}} &
  \multicolumn{1}{c|}{82.26 \scriptsize{± 0.25}} &
  \multicolumn{1}{c|}{67.01 \scriptsize{± 0.39}} &
  {\ul 84.47} \scriptsize{± 0.28} \\
\multicolumn{1}{l|}{Meta-DM + UniSiam + dist$^*$ \citep{metadm_uni}} &
  \multicolumn{1}{c|}{ \ \ \ \ $\texttt{RN18}$ \ \ \ \ } &
  \multicolumn{1}{c|}{ \ \ \ \ $\texttt{Unsup.}$ \ \ \ \ } &
  \multicolumn{1}{c|}{{\ul 65.64} \scriptsize{± 0.36}} &
  \multicolumn{1}{c|}{{\ul 83.97} \scriptsize{± 0.25}} &
  \multicolumn{1}{c|}{{\ul 67.11} \scriptsize{± 0.40}} &
  84.39 \scriptsize{± 0.28} \\ [1pt] \cdashlinelr{1-7} 
  \multicolumn{1}{l|}{MetaOptNet \citep{metaoptnet}} &
  \multicolumn{1}{c|}{$\texttt{RN18}$} &
  \multicolumn{1}{c|}{$\texttt{Sup.}$} &
  \multicolumn{1}{c|}{64.09 \scriptsize{± 0.62}} &
  \multicolumn{1}{c|}{80.00 \scriptsize{± 0.45}} &
  \multicolumn{1}{c|}{65.99 \scriptsize{± 0.72}} &
  81.56 \scriptsize{± 0.53} \\
\multicolumn{1}{l|}{Transductive CNAPS \citep{transdcnaps}} &
  \multicolumn{1}{c|}{$\texttt{RN18}$} &
  \multicolumn{1}{c|}{$\texttt{Sup.}$} &
  \multicolumn{1}{c|}{55.60 \scriptsize{± 0.90}} &
  \multicolumn{1}{c|}{73.10 \scriptsize{± 0.70}} &
  \multicolumn{1}{c|}{65.90 \scriptsize{± 1.10}} &
  81.80 \scriptsize{± 0.70} \\
\multicolumn{1}{l|}{MAML \citep{maml}} &
  \multicolumn{1}{c|}{$\texttt{RN34}$} &
  \multicolumn{1}{c|}{$\texttt{Sup.}$} &
  \multicolumn{1}{c|}{51.46 \scriptsize{± 0.90}} &
  \multicolumn{1}{c|}{65.90 \scriptsize{± 0.79}} &
  \multicolumn{1}{c|}{51.67 \scriptsize{± 1.81}} &
  70.30 \scriptsize{± 1.75} \\
  \multicolumn{1}{l|}{ProtoNet \citep{protonet}} &
  \multicolumn{1}{c|}{$\texttt{RN34}$} &
  \multicolumn{1}{c|}{$\texttt{Sup.}$} &
  \multicolumn{1}{c|}{53.90 \scriptsize{± 0.83}} &
  \multicolumn{1}{c|}{74.65 \scriptsize{± 0.64}} &
  \multicolumn{1}{c|}{51.67 \scriptsize{± 1.81}} &
  70.30 \scriptsize{± 1.75} \\ 
\midrule
\rowcolor[HTML]{E7F5F8} 
\multicolumn{1}{l|}{\cellcolor[HTML]{E7F5F8}\textbf{\ourmethod{} \textbf{(Ours)}}} &
  \multicolumn{1}{c|}{\cellcolor[HTML]{E7F5F8}$\texttt{RN18}$} &
  \multicolumn{1}{c|}{\cellcolor[HTML]{E7F5F8}$\texttt{Unsup.}$} &
  \multicolumn{1}{c|}{\cellcolor[HTML]{E7F5F8}\textbf{75.74} \scriptsize{± 0.62}} &
  \multicolumn{1}{c|}{\cellcolor[HTML]{E7F5F8}\textbf{84.93} \scriptsize{± 0.33}} &
  \multicolumn{1}{c|}{\cellcolor[HTML]{E7F5F8}\textbf{76.44} \scriptsize{± 0.66}} &
  \multicolumn{1}{c}{\cellcolor[HTML]{E7F5F8}\textbf{84.85} \scriptsize{± 0.37}} \\ \midrule
\multicolumn{1}{l|}{SwAV$^\dagger$ \citep{SwAV}} &
  \multicolumn{1}{c|}{$\texttt{RN50}$} &
  \multicolumn{1}{c|}{$\texttt{Unsup.}$} &
  \multicolumn{1}{c|}{63.34 \scriptsize{± 0.42}} &
  \multicolumn{1}{c|}{82.76 \scriptsize{± 0.24}} &
  \multicolumn{1}{c|}{68.02 \scriptsize{± 0.52}} &
  85.93 \scriptsize{± 0.33} \\
\multicolumn{1}{l|}{NNCLR$^\dagger$ \citep{nnclr}} &
  \multicolumn{1}{c|}{$\texttt{RN50}$} &
  \multicolumn{1}{c|}{$\texttt{Unsup.}$} &
  \multicolumn{1}{c|}{65.42 \scriptsize{± 0.44}} &
  \multicolumn{1}{c|}{83.31 \scriptsize{± 0.21}} &
  \multicolumn{1}{c|}{{\ul 69.82} \scriptsize{± 0.54}} &
  86.41 \scriptsize{± 0.31} \\
\multicolumn{1}{l|}{TrainProto \citep{li2021trainable}} &
  \multicolumn{1}{c|}{$\texttt{RN50}$} &
  \multicolumn{1}{c|}{$\texttt{Unsup.}$} &
  \multicolumn{1}{c|}{58.92 \scriptsize{± 0.91}} &
  \multicolumn{1}{c|}{73.94 \scriptsize{± 0.63}} &
  \multicolumn{1}{c|}{-} &
  - \\
\multicolumn{1}{l|}{UBC-FSL \citep{ubc-fsl}} &
  \multicolumn{1}{c|}{$\texttt{RN50}$} &
  \multicolumn{1}{c|}{$\texttt{Unsup.}$} &
  \multicolumn{1}{c|}{56.20 \scriptsize{± 0.60}} &
  \multicolumn{1}{c|}{75.40 \scriptsize{± 0.40}} &
  \multicolumn{1}{c|}{66.60 \scriptsize{± 0.70}} &
  83.10 \scriptsize{± 0.50} \\
\multicolumn{1}{l|}{PDA-Net \citep{pdanet}} &
  \multicolumn{1}{c|}{$\texttt{RN50}$} &
  \multicolumn{1}{c|}{$\texttt{Unsup.}$} &
  \multicolumn{1}{c|}{63.84 \scriptsize{± 0.91}} &
  \multicolumn{1}{c|}{83.11 \scriptsize{± 0.56}} &
  \multicolumn{1}{c|}{69.01 \scriptsize{± 0.93}} &
  84.20 \scriptsize{± 0.69} \\
\multicolumn{1}{l|}{UniSiam + dist \citep{unisiam}} &
  \multicolumn{1}{c|}{$\texttt{RN50}$} &
  \multicolumn{1}{c|}{$\texttt{Unsup.}$} &
  \multicolumn{1}{c|}{65.33 \scriptsize{± 0.36}} &
  \multicolumn{1}{c|}{83.22 \scriptsize{± 0.24}} &
  \multicolumn{1}{c|}{69.60 \scriptsize{± 0.38}} &
  86.51 \scriptsize{± 0.26} \\
\multicolumn{1}{l|}{Meta-DM + UniSiam + dist$^*$ \citep{metadm_uni}} &
  \multicolumn{1}{c|}{$\texttt{RN50}$} &
  \multicolumn{1}{c|}{$\texttt{Unsup.}$} &
  \multicolumn{1}{c|}{{\ul 66.68} \scriptsize{± 0.36}} &
  \multicolumn{1}{c|}{{\ul 85.29} \scriptsize{± 0.23}} &
  \multicolumn{1}{c|}{69.61 \scriptsize{± 0.38}} &
  {\ul 86.53} \scriptsize{± 0.26} \\ \midrule
\rowcolor[HTML]{E7F5F8} 
\multicolumn{1}{l|}{\cellcolor[HTML]{E7F5F8}\textbf{\ourmethod{} \textbf{(Ours)}}} &
  \multicolumn{1}{c|}{\cellcolor[HTML]{E7F5F8}$\texttt{RN50}$} &
  \multicolumn{1}{c|}{\cellcolor[HTML]{E7F5F8}$\texttt{Unsup.}$} &
  \multicolumn{1}{c|}{\cellcolor[HTML]{E7F5F8}\textbf{80.57} \scriptsize{± 0.57}} &
  \multicolumn{1}{c|}{\cellcolor[HTML]{E7F5F8}\textbf{87.82} \scriptsize{± 0.29}} &
  \multicolumn{1}{c|}{\cellcolor[HTML]{E7F5F8}\textbf{81.69} \scriptsize{± 0.61}} &
  \multicolumn{1}{c}{\cellcolor[HTML]{E7F5F8}\textbf{87.86} \scriptsize{± 0.32}} \\ \bottomrule
\end{tabular}%
}
\vspace{-3mm}
\end{table}
\begin{table}[th!]
\centering
\aboverulesep = 0pt
\belowrulesep = 0pt
\renewcommand{\arraystretch}{1.2}
\caption{\small Extended version of Table~\ref{tab:cifar-main}. Accuracies in (\% $\pm$ std) on CIFAR-FS and FC$100$ of different U-FSL baselines. Encoders: \texttt{RN}: ResNet. $^\dagger$: denotes our reproduction. Style: \textbf{best} and \underline{second best}.}
\vspace{-0.1cm}
\label{tab:cifar}
\resizebox{\textwidth}{!}{%
\begin{tabular}{@{}lcccccc@{}}
\toprule
 &
   &
  \multicolumn{1}{c|}{} &
  \multicolumn{2}{c|}{\cellcolor[HTML]{DCE7D9}\textbf{CIFAR-FS}} &
  \multicolumn{2}{c}{\cellcolor[HTML]{FDE5C9}\textbf{FC100}} \\ \midrule
\multicolumn{1}{l|}{\textbf{Method}} &
  \multicolumn{1}{c|}{\ \  \ \textbf{Backbone}\ \  \ } &
  \multicolumn{1}{c|}{\textbf{Setting}} &
  \multicolumn{1}{c|}{\ \  \ \textbf{5-way 1-shot}\ \ \ } &
  \multicolumn{1}{c|}{\ \ \ \textbf{5-way 5-shot}\ \ \ } &
  \multicolumn{1}{c|}{\ \ \ \textbf{5-way 1-shot}\ \ \ } &
  \ \ \ \textbf{5-way 5-shot}\ \ \  \\ \midrule 
\multicolumn{1}{l|}{Meta-GMVAE \citep{meta_gmvae}} &
  \multicolumn{1}{c|}{$\texttt{RN18}$} &
  \multicolumn{1}{c|}{ \ \ \ \ $\texttt{Unsup.}$ \ \ \ \ } &
  \multicolumn{1}{c|}{-} &
  \multicolumn{1}{c|}{-} &
  \multicolumn{1}{c|}{36.30 \scriptsize{± 0.70}} &
  49.70 \scriptsize{± 0.80} \\ 
\multicolumn{1}{l|}{SimCLR \citep{simclr}} &
  \multicolumn{1}{c|}{$\texttt{RN18}$} &
  \multicolumn{1}{c|}{$\texttt{Unsup.}$} &
  \multicolumn{1}{c|}{54.56 \scriptsize{± 0.19}} &
  \multicolumn{1}{c|}{71.19 \scriptsize{± 0.18}} &
  \multicolumn{1}{c|}{36.20 \scriptsize{± 0.70}} &
  49.90 \scriptsize{± 0.70} \\ 
\multicolumn{1}{l|}{MoCo v2 \citep{mocov2}} &
  \multicolumn{1}{c|}{$\texttt{RN18}$} &
  \multicolumn{1}{c|}{$\texttt{Unsup.}$} &
  \multicolumn{1}{c|}{52.73 \scriptsize{± 0.20}} &
  \multicolumn{1}{c|}{67.81 \scriptsize{± 0.19}} &
  \multicolumn{1}{c|}{37.70 \scriptsize{± 0.70}} &
  53.20 \scriptsize{± 0.70} \\ 
\multicolumn{1}{l|}{MoCHi \citep{kalantidis2020hard}} &
  \multicolumn{1}{c|}{$\texttt{RN18}$} &
  \multicolumn{1}{c|}{$\texttt{Unsup.}$} &
  \multicolumn{1}{c|}{50.42 \scriptsize{± 0.22}} &
  \multicolumn{1}{c|}{65.91 \scriptsize{± 0.20}} &
  \multicolumn{1}{c|}{37.51 \scriptsize{± 0.17}} &
  48.95 \scriptsize{± 0.17} \\ 
\multicolumn{1}{l|}{BYOL \citep{byol}} &
  \multicolumn{1}{c|}{$\texttt{RN18}$} &
  \multicolumn{1}{c|}{$\texttt{Unsup.}$} &
  \multicolumn{1}{c|}{51.33 \scriptsize{± 0.21}} &
  \multicolumn{1}{c|}{66.73 \scriptsize{± 0.18}} &
  \multicolumn{1}{c|}{37.20 \scriptsize{± 0.70}} &
  52.80 \scriptsize{± 0.60} \\ 
\multicolumn{1}{l|}{LF2CS \citep{lf2cs}} &
  \multicolumn{1}{c|}{$\texttt{RN18}$} &
  \multicolumn{1}{c|}{$\texttt{Unsup.}$} &
  \multicolumn{1}{c|}{{\ul 55.04} \scriptsize{± 0.72}} &
  \multicolumn{1}{c|}{70.62 \scriptsize{± 0.57}} &
  \multicolumn{1}{c|}{37.20 \scriptsize{± 0.70}} &
  52.80 \scriptsize{± 0.60} \\
\multicolumn{1}{l|}{CUMCA \citep{xu2021unsupervised}} &
  \multicolumn{1}{c|}{$\texttt{RN18}$} &
  \multicolumn{1}{c|}{$\texttt{Unsup.}$} &
  \multicolumn{1}{c|}{50.48 \scriptsize{± 0.12}} &
  \multicolumn{1}{c|}{67.83 \scriptsize{± 0.18}} &
  \multicolumn{1}{c|}{33.00 \scriptsize{± 0.17}} &
  47.41 \scriptsize{± 0.19}\\
\multicolumn{1}{l|}{Barlow Twins \citep{barlowtwins}  \ \  \ \  \ \  \ \ \ \ \ \ \ \  \ \ \ \ } &
  \multicolumn{1}{c|}{$\texttt{RN18}$} &
  \multicolumn{1}{c|}{$\texttt{Unsup.}$} &
  \multicolumn{1}{c|}{-} &
  \multicolumn{1}{c|}{-} &
  \multicolumn{1}{c|}{37.90 \scriptsize{± 0.70}} &
  54.10 \scriptsize{± 0.60} \\ 
\multicolumn{1}{l|}{HMS \citep{hms}} &
  \multicolumn{1}{c|}{$\texttt{RN18}$} &
  \multicolumn{1}{c|}{$\texttt{Unsup.}$} &
  \multicolumn{1}{c|}{54.65 \scriptsize{± 0.20}} &
  \multicolumn{1}{c|}{{\ul 73.70} \scriptsize{± 0.18}} &
  \multicolumn{1}{c|}{37.88 \scriptsize{± 0.16}} &
  53.68 \scriptsize{± 0.18} \\ 
\multicolumn{1}{l|}{Deep Eigenmaps \citep{deepeigenmaps}} &
  \multicolumn{1}{c|}{$\texttt{RN18}$} &
  \multicolumn{1}{c|}{$\texttt{Unsup.}$} &
  \multicolumn{1}{c|}{-} &
  \multicolumn{1}{c|}{-} &
  \multicolumn{1}{c|}{{\ul 39.70} \scriptsize{± 0.70}} &
  {\ul 57.90} \scriptsize{± 0.70} \\ \midrule
\multicolumn{1}{l|}{\cellcolor[HTML]{E7F5F8}\textbf{\ourmethod{} \textbf{(Ours)}}} &
  \multicolumn{1}{c|}{\cellcolor[HTML]{E7F5F8}$\texttt{RN18}$} &
  \multicolumn{1}{c|}{\cellcolor[HTML]{E7F5F8}$\texttt{Unsup.}$} &
  \multicolumn{1}{c|}{\cellcolor[HTML]{E7F5F8}\textbf{70.39} \tiny{± 0.62}} &
  \multicolumn{1}{c|}{\cellcolor[HTML]{E7F5F8}\textbf{81.56} \tiny{± 0.39}} &
  \multicolumn{1}{c|}{\cellcolor[HTML]{E7F5F8}\textbf{45.21} \tiny{± 0.50}} &
  \multicolumn{1}{c}{\cellcolor[HTML]{E7F5F8}\textbf{60.02} \tiny{± 0.43}} \\ \bottomrule
\end{tabular}%
}
\vspace{-3mm}
\end{table}
%

\subsection{In-Domain Setting}
\label{ssec:appendix-in-domain}
\vspace{-0.2cm}
Here, we provide more extensive experimental results on miniImageNet, tieredImageNet, CIFAR-FS and FC$100$ by comparing against additional baselines. Table~\ref{tab:mini-tiered-full} corresponds to an extended version of Table~\ref{tab:mini-tiered} and, similarly, Table~\ref{tab:cifar} is an extended version of Table~\ref{tab:cifar-main}. We assess the performance of \ourmethod{} against a wide variety of methods: from (i) established SSL baselines \citep{simclr,SwAV,byol,kalantidis2020hard,mocov2,barlowtwins,xu2021unsupervised,simsiam,nnclr} to (ii) state-of-the-art U-FSL approaches \citep{Cactus-proto,umtra,prototransfer,meta_gmvae,ubc-fsl,li2021trainable,hms,c3lr,lf2cs,wang2022contrastive,deepeigenmaps,unisiam,psco,samptransfer,metadm_uni}. Furthermore, we compare with a set of supervised baselines \citep{maml,protonet,LEO,ccrot,metaoptnet,transdcnaps}. 


\begin{table}[t]
\centering
\aboverulesep = 0pt
\belowrulesep = 0pt
\renewcommand{\arraystretch}{1.2}
\caption{\small Extended version of Table~\ref{tab:cdfsl}.  Accuracies (in \% $\pm$ std) on miniImageNet → CDFSL. $^\dagger$: denotes our reproduction. Style: \textbf{best} and \underline{second best}.}
\vspace{-0.1cm}
\label{tab:cdfsl-full}
\resizebox{\textwidth}{!}{%
\begin{tabular}{@{}l|cc|cc|cc|cc@{}}
\toprule
\textbf{} &
  \multicolumn{2}{c|}{\cellcolor[HTML]{DCE7D9}\textbf{ChestX}} &
  \multicolumn{2}{c|}{\cellcolor[HTML]{FDE5C9}\textbf{ISIC}} &
  \multicolumn{2}{c|}{\cellcolor[HTML]{ECF4FF}\textbf{EuroSAT}} &
  \multicolumn{2}{c}{\cellcolor[HTML]{dedee0}\textbf{CropDiseases}} \\ \midrule
\multicolumn{1}{l|}{\textbf{Method}} &
  \multicolumn{1}{c|}{\textbf{5 way 5-shot}} &
  \textbf{5 way 20-shot} &
  \multicolumn{1}{c|}{\textbf{5 way 5-shot}} &
  \textbf{5 way 20-shot} &
  \multicolumn{1}{c|}{\textbf{5 way 5-shot}} &
  \textbf{5 way 20-shot} &
  \multicolumn{1}{c|}{\textbf{5 way 5-shot}} &
  \textbf{5 way 20-shot} \\ \midrule
\multicolumn{1}{l|}{ProtoTransfer \citep{prototransfer}} &
  \multicolumn{1}{c|}{26.71 \scriptsize{± 0.46}} &
  33.82 \scriptsize{± 0.48} &
  \multicolumn{1}{c|}{45.19 \scriptsize{± 0.56}} &
  59.07 \scriptsize{± 0.55} &
  \multicolumn{1}{c|}{75.62 \scriptsize{± 0.67}} &
  86.80 \scriptsize{± 0.42} &
  \multicolumn{1}{c|}{86.53 \scriptsize{± 0.56}} &
  95.06 \scriptsize{± 0.32} \\
\multicolumn{1}{l|}{BYOL \citep{byol}} &
  \multicolumn{1}{c|}{26.39 \scriptsize{± 0.43}} &
  30.71 \scriptsize{± 0.47} &
  \multicolumn{1}{c|}{43.09 \scriptsize{± 0.56}} &
  53.76 \scriptsize{± 0.55} &
  \multicolumn{1}{c|}{83.64 \scriptsize{± 0.54}} &
  89.62 \scriptsize{± 0.39} &
  \multicolumn{1}{c|}{92.71 \scriptsize{± 0.47}} &
  96.07 \scriptsize{± 0.33} \\
\multicolumn{1}{l|}{MoCo v2 \citep{mocov2}} &
  \multicolumn{1}{c|}{25.26 \scriptsize{± 0.44}} &
  29.43 \scriptsize{± 0.45} &
  \multicolumn{1}{c|}{42.60 \scriptsize{± 0.55}} &
  52.39 \scriptsize{± 0.49} &
  \multicolumn{1}{c|}{84.15 \scriptsize{± 0.52}} &
  88.92 \scriptsize{± 0.41} &
  \multicolumn{1}{c|}{87.62 \scriptsize{± 0.60}} &
  92.12 \scriptsize{± 0.46} \\
\multicolumn{1}{l|}{SwAV$^\dagger$ \citep{SwAV}} &
  \multicolumn{1}{c|}{25.70 \scriptsize{± 0.28}} &
  30.41 \scriptsize{± 0.25} &
  \multicolumn{1}{c|}{40.69 \scriptsize{± 0.34}} &
  49.03 \scriptsize{± 0.30} &
  \multicolumn{1}{c|}{84.82 \scriptsize{± 0.24}} &
  90.77 \scriptsize{± 0.26} &
  \multicolumn{1}{c|}{ 88.64 \scriptsize{± 0.26}} &
  95.11 \scriptsize{± 0.21} \\
\multicolumn{1}{l|}{SimCLR \citep{simclr}} &
  \multicolumn{1}{c|}{26.36 \scriptsize{± 0.44}} &
  30.82 \scriptsize{± 0.43} &
  \multicolumn{1}{c|}{43.99 \scriptsize{± 0.55}} &
  53.00 \scriptsize{± 0.54} &
  \multicolumn{1}{c|}{82.78 \scriptsize{± 0.56}} &
  89.38 \scriptsize{± 0.40} &
  \multicolumn{1}{c|}{90.29 \scriptsize{± 0.52}} &
  94.03 \scriptsize{± 0.37} \\
\multicolumn{1}{l|}{NNCLR$^\dagger$ \citep{nnclr}} &
  \multicolumn{1}{c|}{25.74 \scriptsize{± 0.41}} &
  29.54 \scriptsize{± 0.45} &
  \multicolumn{1}{c|}{38.85 \scriptsize{± 0.56}} &
  47.82 \scriptsize{± 0.53} &
  \multicolumn{1}{c|}{83.45 \scriptsize{± 0.57}} &
  90.80 \scriptsize{± 0.39} &
  \multicolumn{1}{c|}{ 90.76 \scriptsize{± 0.57}} &
  95.37 \scriptsize{± 0.37} \\
\multicolumn{1}{l|}{C3LR \citep{c3lr}} &
  \multicolumn{1}{c|}{26.00 \scriptsize{± 0.41}} &
  33.39 \scriptsize{± 0.47} &
  \multicolumn{1}{c|}{45.93 \scriptsize{± 0.54}} &
  59.95 \scriptsize{± 0.53} &
  \multicolumn{1}{c|}{80.32 \scriptsize{± 0.65}} &
  88.09 \scriptsize{± 0.45} &
  \multicolumn{1}{c|}{87.90 \scriptsize{± 0.55}} &
  95.38 \scriptsize{± 0.31} \\
\multicolumn{1}{l|}{SAMPTransfer \citep{samptransfer}} &
  \multicolumn{1}{c|}{26.27 \scriptsize{± 0.44}} &
  {\ul 34.15} \scriptsize{± 0.50} &
  \multicolumn{1}{c|}{{\ul 47.60} \scriptsize{± 0.59}} &
  \textbf{61.28} \scriptsize{± 0.56} &
  \multicolumn{1}{c|}{85.55 \scriptsize{± 0.60}} &
  88.52 \scriptsize{± 0.50} &
  \multicolumn{1}{c|}{91.74 \scriptsize{± 0.55}} &
  96.36 \scriptsize{± 0.28} \\
  \multicolumn{1}{l|}{PsCo \citep{psco}} &
  \multicolumn{1}{c|}{24.78 \scriptsize{± 0.23}} &
  27.69 \scriptsize{± 0.23} &
  \multicolumn{1}{c|}{44.00 \scriptsize{± 0.30}} &
  54.59 \scriptsize{± 0.29} &
  \multicolumn{1}{c|}{81.08 \scriptsize{± 0.35}} &
  87.65 \scriptsize{± 0.28} &
  \multicolumn{1}{c|}{88.24 \scriptsize{± 0.31}} &
  94.95 \scriptsize{± 0.18} \\
\multicolumn{1}{l|}{UniSiam + dist \citep{unisiam}} &
  \multicolumn{1}{c|}{ \textbf{28.18} \scriptsize{± 0.45}} &
  \textbf{34.58} \scriptsize{± 0.46} &
  \multicolumn{1}{c|}{45.65 \scriptsize{± 0.58}} &
  56.54 \scriptsize{± 0.5} &
  \multicolumn{1}{c|}{{\ul 86.53} \scriptsize{± 0.47}} &
  {\ul 93.24} \scriptsize{± 0.30} &
  \multicolumn{1}{c|}{{\ul 92.05} \scriptsize{± 0.50}} &
  {\ul 96.83} \scriptsize{± 0.27} \\ 
\multicolumn{1}{l|}{ConFeSS \citep{confess}} &
  \multicolumn{1}{c|}{{\ul 27.09}} &
  {\ul 33.57} &
  \multicolumn{1}{c|}{\textbf{48.85}} &
  {\ul 60.10} &
  \multicolumn{1}{c|}{84.65} &
  90.40 &
  \multicolumn{1}{c|}{88.88} &
  95.34 \\
\multicolumn{1}{l|}{ATA \citep{ata}} &
  \multicolumn{1}{c|}{24.43 \scriptsize{± 0.2}} &
  - &
  \multicolumn{1}{c|}{45.83 \scriptsize{± 0.3}} &
  - &
  \multicolumn{1}{c|}{83.75 \scriptsize{± 0.4}} &
  - &
  \multicolumn{1}{c|}{90.59 \scriptsize{± 0.3}} &
  - \\ 
  \midrule
\rowcolor[HTML]{E7F5F8} 
\multicolumn{1}{l|}{\cellcolor[HTML]{E7F5F8}\textbf{\ourmethod{} \textbf{(Ours)}}} &
  \multicolumn{1}{c|}{\cellcolor[HTML]{E7F5F8}\textbf{28.46} \scriptsize{± 0.23}} &
  \textbf{34.21} \scriptsize{± 0.25} &
  \multicolumn{1}{c|}{\cellcolor[HTML]{E7F5F8}44.48 \scriptsize{± 0.31}} &
  56.89 \scriptsize{± 0.29} &
  \multicolumn{1}{c|}{\cellcolor[HTML]{E7F5F8}\textbf{88.55} \scriptsize{± 0.23}} &
  \textbf{93.92} \scriptsize{± 0.14} &
  \multicolumn{1}{c|}{\cellcolor[HTML]{E7F5F8}\textbf{93.65} \scriptsize{± 0.25}} &
  \multicolumn{1}{c}{\cellcolor[HTML]{E7F5F8}\textbf{97.72} \scriptsize{± 0.13}}\\ \bottomrule
\end{tabular}%
}
\vspace{-10pt}
\end{table}
%

\begin{wraptable}{r}{0.5\columnwidth}
\vspace{-12pt}
\centering
\aboverulesep = 0pt
\belowrulesep = 0pt
\renewcommand{\arraystretch}{1.2}
\caption{\small Accuracies in (\% $\pm$ std) on miniImageNet → CUB. $^\dagger$: denotes our reproduction. Style: \textbf{best} and \underline{second best}.}
\vspace{-0.1cm}
\label{tab:mini to CUB-full}
\resizebox{0.5\textwidth}{!}{%
\begin{tabular}{@{}l|cc@{}}
\toprule
                         & \multicolumn{2}{c}{\cellcolor[HTML]{DCE7D9}\textbf{miniImageNet → CUB}}                  \\ \midrule
\multicolumn{1}{l|}{\textbf{Method}}          & \multicolumn{1}{c|}{\textbf{5-way 1-shot}}                       & \textbf{5-way 5-shot} \\ \midrule
\multicolumn{1}{l|}{Meta-GMVAE \citep{meta_gmvae}}   & \multicolumn{1}{c|}{38.09 \scriptsize{± 0.47}} & 55.65 \scriptsize{± 0.42} \\
\multicolumn{1}{l|}{SimCLR \citep{simclr}}      & \multicolumn{1}{c|}{38.25 \scriptsize{± 0.49}} & 55.89 \scriptsize{± 0.46} \\
\multicolumn{1}{l|}{MoCo v2 \citep{mocov2}}      & \multicolumn{1}{c|}{39.29 \scriptsize{± 0.47}} & 56.49 \scriptsize{± 0.44} \\
\multicolumn{1}{l|}{BYOL \citep{byol}}         & \multicolumn{1}{c|}{40.63 \scriptsize{± 0.46}} & 56.92 \scriptsize{± 0.43} \\
\multicolumn{1}{l|}{SwAV$^\dagger$ \citep{SwAV}}      & \multicolumn{1}{c|}{38.34 \scriptsize{± 0.51}} & 53.94 \scriptsize{± 0.43} \\
\multicolumn{1}{l|}{NNCLR$^\dagger$ \citep{nnclr}}      & \multicolumn{1}{c|}{39.37 \scriptsize{± 0.53}} & 54.78 \scriptsize{± 0.42} \\
\multicolumn{1}{l|}{Barlow Twins \citep{barlowtwins}} & \multicolumn{1}{c|}{40.46 \scriptsize{± 0.47}} & 57.16 \scriptsize{± 0.42} \\
\multicolumn{1}{l|}{Laplacian Eigenmaps \citep{deepeigenmaps}} & \multicolumn{1}{c|}{{\ul 41.08} \scriptsize{± 0.48}}                          & {\ul 58.86} \scriptsize{± 0.45}    \\
\multicolumn{1}{l|}{HMS \citep{hms}}          & \multicolumn{1}{c|}{40.75}        & 58.32        \\ 
\multicolumn{1}{l|}{PsCo   \citep{psco}}      & \multicolumn{1}{c|}{-}            & 57.38   \scriptsize{± 0.44}      \\
\midrule
\rowcolor[HTML]{E7F5F8} 
\multicolumn{1}{l|}{\cellcolor[HTML]{E7F5F8}\textbf{\ourmethod{} \textbf{(Ours)}}} &\multicolumn{1}{c|}{\cellcolor[HTML]{E7F5F8}\textbf{43.45} \scriptsize{± 0.50}} & \multicolumn{1}{c}{\cellcolor[HTML]{E7F5F8}\textbf{59.51} \scriptsize{± 0.46}} \\ \bottomrule
\end{tabular}%
}
\vspace{-12pt}
\end{wraptable}
%
\vspace{-0.2cm}
\subsection{Cross-Domain Setting}
\label{ssec:appendix-cross-domain}
\vspace{-0.2cm}
We also provide an extended version of the experimental results in the miniImageNet → CDFSL cross-domain setting of Table~\ref{tab:cdfsl}, where we compare against additional baselines, as seen in Table~\ref{tab:cdfsl-full}. Additionally, in Table~\ref{tab:mini to CUB-full} we evaluate the performance of \ourmethod{} on the miniImageNet → CUB cross-domain setting. We compare against any existing unsupervised baselines \citep{Cactus-proto, umtra,simclr,SwAV,prototransfer,byol,mocov2,barlowtwins,hms,deepeigenmaps,unisiam,samptransfer,c3lr,psco,meta_gmvae}, for which this more challenging cross-domain experiment has been conducted (to the best of our knowledge).

\section{Complexity Analysis}
\label{sec:appendix-complexity}
\vspace{-0.2cm}
In this section, we analyze the computational and time complexity of \ourmethod{} and compare with different contrastive learning baselines, as summarized in Table~\ref{tab:complexity}. Neither \ourmodule{} nor \ourft{} introduce additional trainable parameters; thus, the total parameter count of \ourmethod{} is on par with standard Siamese architectures and dependent on the backbone configuration. \ourmethod{} utilizes a student-teacher EMA architecture, hence needs to store separate weights for $2$ distinct networks, denoted as ResNet $2 \times$, similar to BYOL \citep{byol}. A batch size of $256$ and $512$ is used for training \ourmethod{} on miniImageNet and tieredImageNet, respectively, which again is standard practice in the U-FSL literature. However, \ourmodule{} artificially enhances the batch, on which the contrastive loss is applied,
to $k+1$ times the size of the original, in effect slightly increasing the training time of \ourmethod{}. \ourft{} also introduces additional calculations in the inference time, which nevertheless results in a negligible increase in terms of the average episode inference time of \ourmethod{}.
\begin{table}[h]
\centering
\renewcommand{\arraystretch}{1.1}
\caption{\small Comparison of the computational complexity of \ourmethod{} with different contrastive SSL approaches in terms of parameter count, training, and inference times. $^\dagger$: denotes our reproduction.}
\label{tab:complexity}
\resizebox{\textwidth}{!}{%
\begin{tabular}{@{}lcccc@{}}
\toprule
\multicolumn{1}{l}{\cellcolor[HTML]{FFFFFF}\textbf{Method}}  & \textbf{Backbone Architecture}       & \textbf{Backbone Parameter Count (M)} & \textbf{Training Time (sec/epoch)}  & \textbf{Inference Time (sec/episode)} \\ \midrule \midrule
\multicolumn{1}{l}{\cellcolor[HTML]{FFFFFF}SwAV$^\dagger$ \citep{SwAV}}    & ResNet-$18 $ $ (1 \times)$ & $11.2$      & $124.51$ & \multicolumn{1}{c}{\cellcolor[HTML]{FFFFFF}$0.213$}  \\
\multicolumn{1}{l}{\cellcolor[HTML]{FFFFFF}NNCLR$^\dagger$ \citep{nnclr}}    & ResNet-$18 $ $(2 \times)$ & $22.4$      & $174.13$  & \multicolumn{1}{c}{\cellcolor[HTML]{FFFFFF}$0.211$}  \\
\multicolumn{1}{l}{\cellcolor[HTML]{FFFFFF}UniSiam \citep{unisiam}} & ResNet-$18 $ $(1 \times)$& $11.2$      & $153.3$7  & \multicolumn{1}{c}{\cellcolor[HTML]{FFFFFF}$0.212$}  \\
\rowcolor[HTML]{E7F5F8} 
\multicolumn{1}{l}{\cellcolor[HTML]{E7F5F8}BECLR (Ours)}   & ResNet-$18 $ $(2 \times)$ & $22.4$      & $190.20$  & \multicolumn{1}{c}{\cellcolor[HTML]{E7F5F8}$0.216$}  \\ \midrule
\multicolumn{1}{l}{\cellcolor[HTML]{FFFFFF}SwAV$^\dagger$ \citep{SwAV}}     & ResNet-$50 $ $(1 \times)$ & $23.5$      & $136.82$  & \multicolumn{1}{c}{\cellcolor[HTML]{FFFFFF}$0.423$}  \\
\multicolumn{1}{l}{\cellcolor[HTML]{FFFFFF}NNCLR$^\dagger$ \citep{nnclr}}   & ResNet-$50 $ $(2 \times)$ & $47.0$      & $182.33$  & \multicolumn{1}{c}{\cellcolor[HTML]{FFFFFF}$0.415$}  \\
\rowcolor[HTML]{FFFFFF} 
\multicolumn{1}{l}{\cellcolor[HTML]{FFFFFF}UniSiam \citep{unisiam}}   & ResNet-$50 $ $(1 \times)$ & $23.5$      & $167.71$  & \multicolumn{1}{c}{\cellcolor[HTML]{FFFFFF}$0.419$}  \\
\rowcolor[HTML]{E7F5F8} 
\multicolumn{1}{l}{\cellcolor[HTML]{E7F5F8}BECLR (Ours)}   &  ResNet-$50 $ $(2 \times)$ & $47.0$      & $280.53$  & \multicolumn{1}{c}{\cellcolor[HTML]{E7F5F8}$0.446$} \\ \bottomrule
\end{tabular}%
}
\end{table}
%

\section{Pseudocode}
\label{sec:appendix-pseudocode}
\vspace{-0.2cm}
This section includes the algorithms for the pretraining methodology of \ourmethod{} and the proposed dynamic clustered memory (\ourmodule{}) in a Pytorch-like pseudocode format. Algorithm~\ref{alg:BECLR} provides an overview of the pretraining stage of \ourmethod{} and is equivalent to Algorithm~\ref{alg:BECLR-ver1}, while Algorithm~\ref{alg:DyCE} describes the two informational paths of \ourmodule{} and is equivalent to Algorithm~\ref{alg:DyCE-ver1}.

\begin{algorithm}[h]
    \scriptsize{}
    \caption{Unsupervised Pretraining of \ourmethod{}: PyTorch-like Pseudocode}\label{alg:BECLR}
    \setstretch{1.1}
    \SetKwInOut{Input}{input}
    \SetKwInOut{Output}{output}
    \SetKwInput{Require}{Require}
	\SetKwInput{Return}{Return}
	\SetKw{Let}{let}
	\SetKwRepeat{Do}{do}{while}
	
	\SetAlgoLined
	\DontPrintSemicolon
	\SetNoFillComment

    \textcolor{darkgreen}{\# \{f, g, h\}\_student: student backbone, projector, and predictor}

    \textcolor{darkgreen}{\# \{f, g\}\_teacher: teacher backbone and projector}

    \textcolor{darkgreen}{\# DyCE\_\{student, teacher\}: our dynamic clustered memory module for student and teacher paths (see Algorithm.~\ref{alg:DyCE})}

    \textcolor{orange}{def} \hskip0.4em\textcolor{cyan}{BECLR}(x): \hfill  \textcolor{darkgreen}{\# x: a random training mini-batch of L samples}

    \hskip2em x = [x1, \hskip0.45em x2] = [aug1(x), \hskip0.45em aug2(x)] \hfill  \textcolor{darkgreen}{\# concatenate the two augmented views of x} 
    
    \hskip2em z\_s = h\_student( g\_student( f\_student( mask(x)))) \hfill  \textcolor{darkgreen}{\# (2B $\times$ d): extract student representations }
    
    \hskip2em z\_t = g\_teacher( f\_teacher(x)).detach() \hfill  \textcolor{darkgreen}{\# (2B $\times$ d): extract teacher representations }
    
    \hskip2em z\_s, z\_t = \textcolor{cyan}{DyCE\_student}(z\_s), \textcolor{cyan}{DyCE\_teacher}(z\_t) \hfill \textcolor{darkgreen}{\# update memory via optimal transport \& compute enhanced batch (2B(k+1) $\times$ d)}

    \hskip2em loss\_pos = - (z\_s $*$ z\_t).sum(dim=1).mean() \hfill  \textcolor{darkgreen}{\# compute positive loss term } 

    \hskip2em loss = loss\_pos + (matmul(z\_s, z\_t.T) $*$ mask).div(temp).exp().sum(dim=1)).div(n\_neg).mean().log() \hfill  \textcolor{darkgreen}{\# compute final loss term } 

     \hskip2em loss.backward(), \hskip0.45em momentum\_update( student.parameters, teacher.parameters)\hfill  \textcolor{darkgreen}{\# update student and teacher parameters } 
    

\end{algorithm}

\begin{algorithm}[h]
    \scriptsize{}
    \caption{ Dynamic Clustered Memory (\ourmodule{}): PyTorch-like Pseudocode}\label{alg:DyCE}
    \setstretch{1.03}
    \SetKwInOut{Input}{Input}
    \SetKwInOut{Output}{Output}
    \SetKwInput{Require}{Require}
    \SetKwInput{Return}{Return}
    \SetKw{Let}{let}
    \SetKwRepeat{Do}{do}{while}
    \SetAlgoLined
    \DontPrintSemicolon
    \SetNoFillComment
 
    \textcolor{darkgreen}{\# z: batch representations (2B$\times$d)}

    \textcolor{darkgreen}{\# self.memory: memory embedding space (M$\times$d)}

    \textcolor{darkgreen}{\# self.prototypes: memory partition prototypes (P$\times$d)}
    
    \textcolor{orange}{def} \hskip0.2em \textcolor{cyan}{DyCE}(self, z):

    \hskip1em \textcolor{orange}{if} \hskip0.45em self.memory.shape[0] == M:

    \hskip2em \textcolor{darkgreen}{\# - - - - - \textbf{Path I}: Top-k NNs Selection and Batch Enhancement - - - - - }

    \hskip2em \textcolor{orange}{if} \hskip0.45em epoch $\geq$ epoch\_thr

    \hskip3em batch\_prototypes = assign\_prototypes(z, self.prototypes) \hfill \textcolor{darkgreen}{\# (2B$\times$d): find nearest memory prototype for each batch embedding}

    \hskip3em y\_mem = topk(self.memory, z, batch\_prototypes) \hfill \textcolor{darkgreen}{\# (2Bk$\times$d): find top-k NNs, from memory partition of nearest prototype}
    
    \hskip3em z = [z, y\_mem]  \hfill \textcolor{darkgreen}{\# (2B(k+1)$\times$d): concatenate batch and memory representations to create the final enhanced batch}

    \hskip2em \textcolor{darkgreen}{\# - - - - - \textbf{Path II}: Iterative Memory Updating - - - - - }

    \hskip2em opt\_plan = sinkhorn( D(z, self.prototypes)) \hfill \textcolor{darkgreen}{\# get optimal assignments between batch embeddings and prototypes (Solve Eq.~\ref{eqn:opt-transp})}

    \hskip2em self.update(z, opt\_plan) \hfill \textcolor{darkgreen}{\# add latest batch to memory and update memory partitions and prototypes, using the optimal assignments}

    \hskip2em self.dequeue() \hfill \textcolor{darkgreen}{\# discard the 2B oldest memory embeddings}

    \hskip1em \textcolor{orange}{else}:   

    \hskip2em self.enqueue(z) \hfill \textcolor{darkgreen}{\# simply store latest batch until the memory is full for the first time}
    
    \hskip1em \textcolor{orange}{return} \hskip0.35em z
\end{algorithm}
%

\vspace{-0.2cm}
\section{In-Depth Comparison with PsCo}
\label{ssec:appendix-psco}
\vspace{-0.2cm}
In this section we perform an comparative analysis of \ourmethod{} with PsCo \citep{psco}, in terms of their motivation, similarities, discrepancies, and performance. 

\subsection{Motivation and Design Choices}
\ourmethod{} and PsCo indeed share some similarities, in that both methods utilize a student-teacher momentum architecture, a memory module of past representations, some form of contrastive loss, and optimal transport (even though for different purposes). Note that none of these components are unique to neither PsCo nor $\texttt{BECLR}$, but can be found in the overall U-FSL and SSL literature \citep{moco,nnclr,unisiam,wang2022contrastive,hms}. 

Let us now expand on their discrepancies and the unique aspects of \ourmethod{}: (i) PsCo is based on meta learning, constructing few-shot classification (i.e., $N$-way $K$-shot) tasks, during meta-training and relies on fine-tuning to rapidly adapt to novel tasks during meta-testing. In contrast, \ourmethod{} is a contrastive framework based on metric/transfer learning, which focuses on representation quality and relies on \ourft{} for transferring to novel tasks (no few-shot tasks during pretraining, no fine-tuning during inference/testing). (ii) PsCo utilizes a simple FIFO memory queue and is oblivious to class-level information, while \ourmethod{} maintains a clustered highly-separable (as seen in Fig.~\ref{fig:mem-evol}) latent space in \ourmodule{}, which, after an adaptation period, is used for sampling meaningful positives. (iii) PsCo applies optimal transport for creating pseudolabels (directly from the unstructured memory queue) for $N*K$ support embeddings, in turn used as a supervisory signal to enforce consistency between support (drawn from teacher) and query (student) embeddings of the created pseudolabeled task. In stark contrast, \ourmethod{} artificially enhances the batch with additional positives and applies an instance-level contrastive loss to enforce consistency between the original and enhanced (additional) positive pairs. After each training iteration, optimal transport is applied to update the stored clusters within \ourmodule{} in an equipartitioned fashion with embeddings from the current batch. (iv) Finally, \ourmethod{} also incorporates optimal transport (in \ourft{}) to align the distributions between support and query sets, during inference, which does not share similarity with the end-to-end pipeline of PsCo.

\begin{wraptable}{r}{0.53\columnwidth}
\vspace{-12pt}
\centering
\aboverulesep = 0pt
\belowrulesep = 0pt
\renewcommand{\arraystretch}{1.2}
\caption{\small Accuracies in (\% $\pm$ std) on miniImageNet. $^\dagger$: denotes our reproduction. Style: \textbf{best} and \underline{second best}.}
\label{tab:psco-study}
\resizebox{0.53\textwidth}{!}{%
\begin{tabular}{@{}lccc@{}}
\toprule
\textbf{Method}  & \textbf{Backbone} & \textbf{5 way 1 shot} & \textbf{5 way 5 shot} \\ \midrule \midrule
\texttt{PsCo} \citep{psco}             & \texttt{Conv5}             & 46.70 ± 0.42          & 63.26 ± 0.37          \\
\texttt{PsCo$^\dagger$} \citep{psco}              & \texttt{RN18}         & 47.24 ± 0.46          & 65.48 ± 0.38          \\
\texttt{PsCo+$^\dagger$} \citep{psco}             & \texttt{RN18}         & 47.86 ± 0.44          & 65.95 ± 0.37          \\
\texttt{PsCo++$^\dagger$} \citep{psco}            & \texttt{RN18}         & 47.58 ± 0.45          & 65.74 ± 0.38          \\ \midrule
\texttt{PsCo} w/ \ourft$^\dagger$ \citep{psco}      & \texttt{RN18}         & 52.89 ± 0.61          & 67.42 ± 0.51          \\
\texttt{PsCo+} w/ \ourft{}$^\dagger$ \citep{psco}     & \texttt{RN18}         & 54.43 ± 0.59          & 68.31 ± 0.52          \\
\texttt{PsCo++} w/ \ourft{}$^\dagger$ \citep{psco}    & \texttt{RN18}         & 54.35 ± 0.60          & 68.43 ± 0.52          \\ \midrule
\texttt{BECLR} (Ours)           & \texttt{RN18}         & \textbf{75.74 ± 0.62} & \textbf{84.93 ± 0.33} \\
\texttt{BECLR-} (Ours)           & \texttt{RN18}         & 74.37 ± 0.61          & 84.19 ± 0.31          \\
\texttt{BECLR--}  (Ours)         & \texttt{RN18}         & 73.65 ± 0.61          & 83.63 ± 0.31          \\ \midrule
\texttt{BECLR} w/o \ourft{} (Ours)    & \texttt{RN18}         & 66.14 ± 0.43          & 84.32 ± 0.27          \\
\texttt{BECLR-} w/o \ourft{} (Ours)  & \texttt{RN18}         & 65.26 ± 0.41          & 83.68 ± 0.25          \\
\texttt{BECLR--} w/o \ourft{} (Ours)  & \texttt{RN18}         & 64.68 ± 0.44          & 83.45 ± 0.26          \\ \bottomrule
\end{tabular}%
}
\vspace{-12pt}
\end{wraptable}
\subsection{Performance and Robustness}

\textcolor{black}{We conduct an additional experiment, summarized in Table~\ref{tab:psco-study}, to study the impact of adding components of \ourmethod{} (symmetric loss, masking, \ourft{}) to PsCo and the robustness of \ourmethod{}. For the purposes of this experiment, we reproduced PsCo on a deeper ResNet-18 backbone, ensuring fairness in our comparisons. Next, we modified its loss function to be symmetric (allowing both augmentations of $X$ to pass through both branches) similar to \ourmethod{}, (this is denoted as $\texttt{PsCo+}$). Next, we also add patch-wise masking to PsCo (denoted as $\texttt{PsCo++}$) and as an additional step we have also added our novel module $\texttt{OpTA}$ on top of all these models (\texttt{PsCo} w/ \ourft{}), during inference. We observe that neither the symmetric loss nor masking offer meaningful improvements in the performance of PsCo. On the contrary, $\texttt{OpTA}$ yields a significant performance boost (up to $6.5\%$ in the $1$-shot setting, in which the sample bias is most severe). This corroborates our claim that our proposed $\texttt{OpTA}$ should be considered as an add-on module to every (U-)FSL approach out there.} 
    
\textcolor{black}{As an additional robustness study we take the opposite steps for \ourmethod{}, first removing the patch-wise masking (denoted as \texttt{BECLR-}) and then also the symmetric loss (denoted as \texttt{BECLR--}), noticing that both slightly degrade the performance, as seen in Table~\ref{tab:psco-study}, which confirms our design choices to include them. Similarly, we also remove \ourft{} from \ourmethod{}'s inference stage. The most important takeaway here is that the most degraded version of $\texttt{BECLR}$ (\texttt{BECLR--} w/o \ourft{}) still outperforms the best enhanced version of PsCo ( \texttt{PsCo++ }w/ \ourft{}), which we believe offers an additional perspective on the advantages of adopting \ourmethod{} over PsCo. }

\subsection{Computational Complexity}

Finally, we also compare \ourmethod{} and PsCo in terms of model size (total number of parameters in M) and inference times (in sec/episode), when using the same ResNet-18 backbone architecture. The results are summarized in Table~\ref{tab:infer-time}. We notice that \ourmethod{}'s total parameter count is, in fact, lower than that of PsCo. Regarding inference time, \ourmethod{} is slightly slower in comparison, but this difference is negligible in real-time inference scenarios. 
\begin{table}[t]
\centering
\caption{\small Comparison of the computational complexity of \ourmethod{} with PsCo \citep{psco} in terms of total model parameter count and inference times.}
\label{tab:infer-time}
\resizebox{0.9\textwidth}{!}{%
\begin{tabular}{@{}lccc@{}}
\toprule
\textbf{Method} & \textbf{Backbone} & \textbf{Model Total Parameter Count (M)} & \textbf{Inference Time (sec/episode)} \\ \midrule \midrule
PsCo \citep{psco}            & \texttt{RN18}         & 30.766                        & 0.082                                 \\
BECLR (Ours)          & \texttt{RN18}         & 24.199                        & 0.216                                 \\ \bottomrule
\end{tabular}%
}
\end{table}
%

\vspace{-0.2cm}
\section{Comparison with Supervised FSL}
\label{ssec:appendix-sup-fsl}
\vspace{-0.2cm}
\textcolor{black}{In this work we have focused on unsupervised few-shot learning and demonstrated how \ourmethod{} sets a new state-of-the-art in this exciting space. In this section, as an additional study, we are also comparing \ourmethod{} with a group of most recent supervised FSL baselines, which also have access to the base-class labels in the pretraining stage. }

\subsection{Performance Evaluation}
\textcolor{black}{In Table~\ref{tab:superivised} we compare \ourmethod{} with seven recent baselines from the supervised FSL state-of-the-art \citep{metaoptnet,transdcnaps,he2022attribute,hiller2022rethinking,bendou2022easy,singh2022transductive,hu2023adaptive}, in terms of their in-domain performance on miniImageNet and tieredImageNet. As can be seen, some of these supervised methods do outperform \ourmethod{}, yet these baselines are heavily engineered towards the target dataset in the in-domain setting (where the target classes still originate from the pretraining dataset). Another interesting observation here is that it turns out the top performing supervised baselines are all \emph{transductive} methodologies (i.e., learn from both labeled and unlabeled data at the same time), and such a transductive episodic pretraining cannot be established in a fully unsupervised pretraining strategy as in \ourmethod{} Notice that \ourmethod{} can even outperform recent inductive supervised FSL approaches, \textit{even without access to base-class labels} during pretraining.}

\begin{table}[h!]
\centering
\aboverulesep = 0pt
\belowrulesep = 0pt
\renewcommand{\arraystretch}{1.3}
\caption{\small \textcolor{black}{Accuracies (in \% $\pm$ std) on miniImageNet and tieredImageNet compared against supervised FSL baselines. Pretrainig Setting: unsupervised (\texttt{Unsup.}) and supervised (\texttt{Sup.}) pretraining. Approach: \texttt{Ind.}: inductive setting, \texttt{Transd.}: transductive setting. Style: \textbf{best} and \underline{second best}.}}
\label{tab:superivised}
\resizebox{\textwidth}{!}{%
\begin{tabular}{@{}lcccc|cc@{}}
\toprule

               &         &              & \multicolumn{2}{c|}{\cellcolor[HTML]{DCE7D9}\textbf{miniImageNet}} & \multicolumn{2}{c}{\cellcolor[HTML]{FDE5C9}\textbf{tieredImageNet}}   \\ \midrule
\multicolumn{1}{l}{\cellcolor[HTML]{FFFFFF}\textbf{Method}}         & \textbf{Setting} & \textbf{Approach}     & \textbf{5 way 1 shot}    & \textbf{5 way 5 shot}   & \textbf{5 way 1 shot} & \textbf{5 way 5 shot}          \\ \midrule
\rowcolor[HTML]{E7F5F8} 
\multicolumn{1}{l}{\cellcolor[HTML]{E7F5F8}\textbf{\ourmethod{} \textbf{(Ours)}}} & \multicolumn{1}{c}{\cellcolor[HTML]{E7F5F8}\texttt{Unsup.}} & \multicolumn{1}{c}{\cellcolor[HTML]{E7F5F8}\texttt{Ind.}}    & \multicolumn{1}{c}{\cellcolor[HTML]{E7F5F8}\textbf{80.57 \scriptsize{± 0.57}}} & \multicolumn{1}{c|}{\cellcolor[HTML]{E7F5F8}\textbf{87.82 \scriptsize{± 0.29}}} & \multicolumn{1}{c}{\cellcolor[HTML]{E7F5F8}\textbf{81.69 \scriptsize{± 0.61}}} & \multicolumn{1}{c}{\cellcolor[HTML]{E7F5F8}87.86 \scriptsize{± 0.32}}          \\ [1pt] \cdashlinelr{1-7} 
\multicolumn{1}{l}{\cellcolor[HTML]{FFFFFF}MetaOptNet \citep{metaoptnet}}     & \texttt{Sup.}    & \texttt{Ind.}    & 64.09 \scriptsize{± 0.62}    & 80.00 \scriptsize{± 0.45}   & 65.99 \scriptsize{± 0.72} & 81.56 \scriptsize{± 0.53}          \\
\multicolumn{1}{l}{\cellcolor[HTML]{FFFFFF}HCTransformers \citep{he2022attribute}} & \texttt{Sup.}    & \texttt{Ind.}    & {\ul 74.74} \scriptsize{± 0.17}    & 85.66 \scriptsize{± 0.10}   & {\ul 79.67} \scriptsize{± 0.20} & {\ul 89.27} \scriptsize{± 0.13}          \\
\multicolumn{1}{l}{\cellcolor[HTML]{FFFFFF}FewTURE \citep{hiller2022rethinking}}        & \texttt{Sup.}    & \texttt{Ind.}    & 72.40 \scriptsize{± 0.78}    & {\ul 86.38} \scriptsize{± 0.49}   & 76.32 \scriptsize{± 0.87} & \textbf{89.96 \scriptsize{± 0.55}} \\
\multicolumn{1}{l}{\cellcolor[HTML]{FFFFFF}EASY (inductive) \citep{bendou2022easy}}          & \texttt{Sup.}    & \texttt{Ind.}    & 70.63 \scriptsize{± 0.20}    & 86.28 \scriptsize{± 0.12}   & 74.31 \scriptsize{± 0.22} & 87.86 \scriptsize{± 0.15}          \\ \midrule
\multicolumn{1}{l}{\cellcolor[HTML]{FFFFFF}EASY (transductive) \citep{bendou2022easy}}         & \texttt{Sup.}    & \texttt{Transd.} & 82.31 \scriptsize{± 0.24}    & 88.57 \scriptsize{± 0.12}   & 83.98 \scriptsize{± 0.24} & 89.26 \scriptsize{± 0.14}          \\
\multicolumn{1}{l}{\cellcolor[HTML]{FFFFFF}Transductive CNAPS \citep{transdcnaps}} & \texttt{Sup.}   & \texttt{Transd.} & 55.60 \scriptsize{± 0.90}          & 73.10 \scriptsize{± 0.70}          & 65.90 \scriptsize{± 1.10}          & 81.80 \scriptsize{± 0.70}          \\
\multicolumn{1}{l}{\cellcolor[HTML]{FFFFFF}BAVARDAGE \citep{hu2023adaptive}}      & \texttt{Sup.}    & \texttt{Transd.} & {\ul 84.80} \scriptsize{± 0.25}    & {\ul 91.65} \scriptsize{± 0.10}   & {\ul 85.20} \scriptsize{± 0.25} & {\ul 90.41} \scriptsize{± 0.14}          \\
\multicolumn{1}{l}{\cellcolor[HTML]{FFFFFF}TRIDENT \citep{singh2022transductive}}            & \texttt{Sup.}   & \texttt{Transd.} & \textbf{86.11 \scriptsize{± 0.59}} & \textbf{95.95 \scriptsize{± 0.28}} & \textbf{86.97 \scriptsize{± 0.50}} & \textbf{96.57 \scriptsize{± 0.17}} \\ \bottomrule
\end{tabular}%
}
\vspace{-10pt}
\end{table}

\subsection{Additional Intuition}
\textcolor{black}{Let us now try to provide some high-level intuition as to why \ourmethod{} can outperform such supervised baselines. We argue that self-supervised pretraining helps generalization to the unseen classes, whereas supervised training heavily tailors the model towards the pretraining classes. This is also evidenced by their optimization objectives. In particular, supervised pretraining maximizes the mutual information $I(\bm{Z}, \bm{y})$ between representations $\bm{Z}$ and base-class labels $\bm{y}$, whereas \ourmethod{} maximizes the mutual information $I(\bm{Z_1}, \bm{Z_2})$ between different augmented views $\bm{Z_1}, \bm{Z_2}$ of the input images $\bm{X}$, which is a lower bound of the mutual information $I(\bm{Z}, \bm{X})$ between representations $\bm{Z}$ and raw data/ images $\bm{X}$. As such, \ourmethod{} \emph{potentially} has a higher capacity to learn more discriminative and generalizable features. By the way, this is not the case only for $\texttt{BECLR}$, many other unsupervised FSL approaches report similar behavior, e.g. \citep{pdanet,unisiam,metadm_uni}. Next to that, pure self-supervised learning approaches also report a similar observation where they can outperform supervised counterparts due to better generalization to different downstream tasks, e.g. \citep{henaff2021efficient,zhang2022dino}. }

\textcolor{black}{Another notable reason why \ourmethod{} can outperform some of its supervised counterparts is our novel module \ourft{} specifically designed to address sample bias, a problem that both supervised and unsupervised FSL approaches suffer from and typically overlook. So comes our claim that the proposed \ourft{} should becomes an integral part of all (U-)FSL approaches, especially in low-shot scenarios where FSL approaches suffer from samples bias the most. }


\end{document}